\documentclass[10pt,twocolumn,letterpaper]{article}

\usepackage[pagenumbers]{wacv} 

\usepackage{indentfirst}
\usepackage[dvipsnames]{xcolor}

\usepackage[ruled,vlined]{algorithm2e}
\definecolor{wacvblue}{rgb}{0.21,0.49,0.74}
\usepackage[pagebackref,breaklinks,colorlinks,allcolors=wacvblue]{hyperref}

\def\wacvPaperID{131} 
\def\confName{WACV}
\def\confYear{2026}

\usepackage{times}
\usepackage{epsfig}
\usepackage{graphicx}
\usepackage{amsmath}
\usepackage{amssymb}
\usepackage{algpseudocode}
\usepackage{float}
\usepackage{color}
\usepackage{colortbl}
\usepackage{rotating}
\usepackage{times}
\usepackage{epsfig}
\usepackage{enumitem}
\usepackage{makecell}
\usepackage{lineno}
\usepackage{tabularx}
\usepackage{xcolor}
\usepackage{multirow}
\usepackage{graphicx}
\usepackage{hhline}
\usepackage{textcomp,booktabs}
\usepackage{caption}
\usepackage{stmaryrd}
\usepackage[mathscr]{eucal}
\usepackage{lineno}
\usepackage[export]{adjustbox}
\usepackage{comment}
\usepackage{multicol,lipsum}
\usepackage{amsfonts}
\usepackage{arydshln}
\usepackage{marvosym}  

\usepackage{url}
\usepackage{tabularx}

\usepackage{stfloats}

\def\1{mathbb{1}}

\def\0{{\bf 0}}
\def\1{{\bf 1}}

\def\RB{{\mathbb R}}

\definecolor{purple}{rgb}{0.56,0.27,0.68}
\definecolor{red}{rgb}{0.95,0.4,0.4}
\definecolor{purered}{rgb}{1,0,0}
\definecolor{blue}{rgb}{0.4,0.4,0.95}
\definecolor{darkblue}{rgb}{0,0,0.8}
\definecolor{lightblue}{rgb}{127,153,240}
\definecolor{grey}{rgb}{0.6,0.6,0.6}
\definecolor{col1}{RGB}{232, 161, 148}
\definecolor{col11}{RGB}{255, 228, 228}
\definecolor{col2}{RGB}{148, 187, 232}
\definecolor{col33}{RGB}{206, 239, 255}
\definecolor{col3}{RGB}{233, 255, 245}
\definecolor{lightgrey}{rgb}{0.85,0.85,0.85}
\definecolor{lightlightgrey}{rgb}{0.9,0.9,0.9}
\definecolor{verylightBG}{rgb}{0.9,0.99,0.99}
\definecolor{darkgreen}{rgb}{0., 0.85, 0.5}

\graphicspath{{./figures/}}   

\title{Roadside Monocular 3D Detection Prompted by 2D Detection}

\author{Yechi Ma$^{1, 2}$, Wei Hua$^2$, Yanan Li$^2$, Shu Kong$^{3, 4, \text{\Letter}}$ \\ \\
{$^1$Zhejiang University, $^2$Zhejiang Lab, $^3$University of Macau, $^4$Institute of Collaborative Innovation}\\
{\em website and code: \url{https://github.com/mayechi/Pro3D}}
  \vspace{-2mm}
}

\begin{document}


\twocolumn[
{%
\maketitle
\vspace{-3mm}
\centering
\includegraphics[width=1.0\linewidth, height=5.5cm]{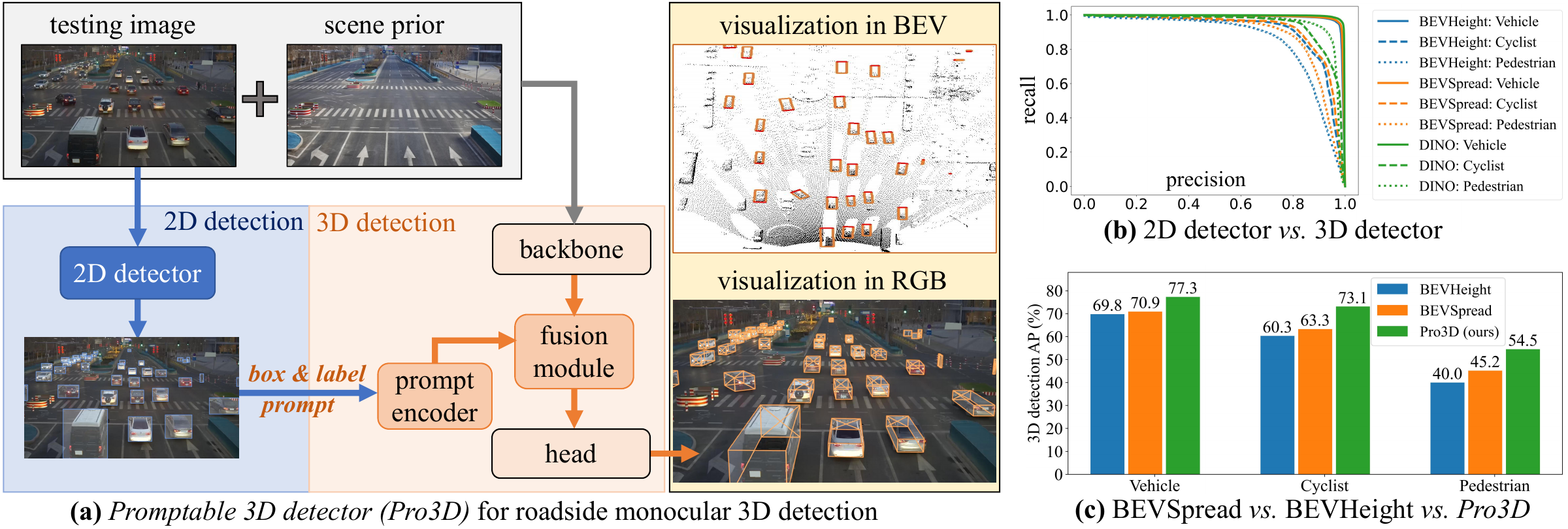}
\vspace{-7mm}
\captionof{figure}{\small
{\bf A summary of our work.}
{\bf (a)} To approach roadside monocular 3D detection, we introduce a novel detector design \emph{Promptable 3D detector (Pro3D)} that exploits 2D detections to prompt the 3D detector for 3D prediction.
Pro3D can exploit any 2D detector and 3D detector with minimal modificaitons.
{\bf (b)} 
The Pro3D design is motivated by the observation that a simple 2D detector DINO~\cite{zhang2022dino} significantly outperforms state-of-the-art roadside monocular 3D detectors (e.g., BEVHeight~\cite{yang2023BEVHeight} and BEVSpread~\cite{wang2024bevspread}) w.r.t 2D metrics (on the DAIR-V2X-I dataset).
This implies that 2D detection is an ``easier'' task than monocular 3D detection -- 
with a trained 2D detector, training 3D detector can be ``simplified'' by learning to lift 2D detections to the 3D space.
Moreover, as roadside camera pose is fixed, we derive a backbround image as scene prior, incorporating which remarkably boosts 3D detection performance.
{\bf (c)} As a summary of results, Pro3D significantly outperforms prior works on the DAIR-V2X-I  benchmark.
Refer to Table~\ref{tab:benchmarking-results(0.5,0.25,0.25)} 
and \ref{tab:benchmarking-results of Rope3d(0.7,0.5,0.5)} for comprehensive results.
}
\label{fig:overview}
\vspace{4mm}
}
]

\begin{abstract}
\noindent
Roadside monocular 3D detection requires detecting objects of predefined classes in an RGB frame and predicting their 3D attributes, such as bird's-eye-view (BEV) locations. It has broad applications in traffic control, vehicle-vehicle communication, and vehicle-infrastructure cooperative perception. To address this task, we introduce Promptable 3D Detector (Pro3D), a novel detector design that leverages 2D detections as prompts. We build our Pro3D upon two key insights. First, compared to a typical 3D detector, a 2D detector is ``easier'' to train due to fewer loss terms and performs significantly better at localizing objects w.r.t 2D metrics. Second, once 2D detections precisely locate objects in the image, a 3D detector can focus on lifting these detections into 3D BEV, especially when fixed camera pose or scene geometry provide an informative prior. To encode and incorporate 2D detections, we explore three methods: (a) concatenating features from both 2D and 3D detectors, (b) attentively fusing 2D and 3D detector features, and (c) encoding properties of predicted 2D bounding boxes  \{$x$, $y$, width, height, label\} and attentively fusing them with the 3D detector feature. Interestingly, the third method significantly outperforms the others, underscoring the effectiveness of 2D detections as prompts that offer precise object targets and allow the 3D detector to focus on lifting them into 3D. Pro3D is adaptable for use with a wide range of 2D and 3D detectors with minimal modifications. Comprehensive experiments demonstrate that our Pro3D significantly enhances existing methods, achieving state-of-the-art results on two contemporary benchmarks.

\end{abstract}

\vspace{-3mm}

\section{Introduction}

Roadside monocular 3D detection is a practical yet challenging task that requires, over a single RGB frame,
detecting objects of predefined classes (e.g., vehicle and pedestrian) and predicting their 3D properties such as the locations, shape, and orientations in bird's-eye-view (BEV)~\cite{yang2023BEVHeight}. 
It has broad applications~\cite{yu2022dair, ye2022rope3d} in vehicle-vehicle communication \cite{chen2019cooper, li2022v2x}, 
vehicle-infrastructure cooperative perception \cite{arnold2020cooperative}, and intelligent traffic control \cite{rauch2012car2x, wang2020v2vnet}.

{\bf Status quo.}
Roadside monocular 3D detection is an ill-posed problem as it requires inferring 3D information from 2D cues.
Fortunately, camera pose is fixed on roadside infrastructure, making it possible to learn from data to infer 3D properties from a 2D image.
To foster the research of roadside monocular 3D detection,
the community has contributed large-scale benchmark datasets, such as DAIR-V2X-I \cite{yu2022dair} and Rope3D \cite{ye2022rope3d}.
Existing methods train sophisticated deep neural networks in an end-to-end fashion~\cite{li2022bevformer, li2023BEVDepth, yang2023BEVHeight, wang2024bevspread}. 
For monocular 3D detection, a notorious difficulty is precisely estimating depth for objects~\cite{welchman2004human, rushton2009observers, gupta2023far3det}.
To mitigate this difficulty,
prior works propose to convert depth estimation to object height estimation and exploits camera parameters to infer object depth~\cite{yang2023BEVHeight}, or consider scene geometry to carefully pool features for better 3D prediction~\cite{wang2024bevspread}.

{\bf Motivation.}
We observe that the state-of-the-art roadside 3D detectors
BEVHeight~\cite{yang2023BEVHeight, yang2023bevheight++} and BEVSpead~\cite{wang2024bevspread} significantly {\em underperform} 2D detectors (e.g.,  DINO~\cite{zhang2022dino}) w.r.t 2D detection metrics,
as shown by the precision-recall curves on the DAIR-V2X-I dataset~\cite{yu2022dair} in  Fig.~\ref{fig:overview}b. 
Here, we train the 2D detector DINO~\cite{zhang2022dino} on this dataset, using the generated 2D bounding boxes by projecting 3D cuboid annotations onto the image plane.
To evaluate BEVHeight w.r.t 2D detection metrics, we project its 3D detections onto the image plane as its 2D detections.
This comparison has several important implications: 
(1) 2D detection is an ``easier'' task than monocular 3D detection, and
(2) the fixed camera pose or scene geometry provides an informative prior that greatly simplifies 2D detection.
The results motivate us to exploit 2D detections and  scene priors to facilitate 3D detector training,
leading to our novel detector design (Fig.~\ref{fig:overview}a), \emph{Promptable 3D Detector (Pro3D)}.

{\bf Technical insights.} 
Training a 3D detector involves optimizing multiple objectives \cite{peri2022towards}, such as 2D box coordinate regression, depth estimation, orientation regression, and object classification.
Intuitively, optimizing them all together might cause the missing detection of hard objects as an attempt to handle the difficulty in regressing coordinates, depth and orientation. 
In contrast, training a 2D detector is simpler that incorporates fewer loss terms, i.e., only a 2D box coordinate regression loss and a classification loss.
This observation leads to our core insight that exploits a 2D detector to facilitate 3D detector training (Fig.~\ref{fig:overview}a).
Intuitively, 2D detections pinpoint the target objects in the image and the 3D detector can be thought of as learning to lift them to 3D.
From a contemporary view, our approach uses a 2D detector to prompt 3D detector, hence we name our framework \emph{Promptable 3D Detector (Pro3D)}.
Moreover, 
as the camera is fixed in roadside perception,
we derive an ``empty'' scene background as a scene prior from training frames (Fig.~\ref{fig:background}).
We incorporate the scene prior as a part of input to the 3D detector.
Extensive experiments validate the effectiveness of the scene prior, and importantly, the design of Pro3D.

{\bf Contributions.} We make three contributions.
\begin{enumerate}
    \item We introduce a novel detector design \emph{Pro3D}, which serves as a general framework for roadside monocular 3D detection (Fig.~\ref{fig:overview}a).
Pro3D is adaptable with any 2D detectors and roaside 3D detectors: it produces 2D detections, then uses them as prompts to help a 3D detector make better 3D detections. We develop multiple methods to exploit the 2D detector and derive the simple yet effective approach -- attentively fusing 2D detections.

    \item To better solve roadside monocular 3D detection, we present an embarrassingly simple method to obtain a scene prior: detecting and removing objects in frames, and averaging the pixels across frames towards an ``empty'' scene background as the scene prior.
    Incorporating the scene prior greatly enhances 3D detection.

    \item We evaluate our Pro3D with different 2D detectors and 3D detectors through extensive experiments and ablation studies, showing that Pro3D significantly improves over existing roadside 3D detectors and achieves the state-of-the-art on two recently released benchmark datasets.
\end{enumerate}




\section{Related Work}

\textbf{Roadside monocular 3D detection} can enlarg the range of perception with a camera hung on a high infrastructure~\cite{yang2023BEVHeight}, serving as an important component in smart transportation and smart city by
To study this problem, the community established large-scale datasets such as DAIR-V2X-I \cite{yu2022dair} and Rope3D \cite{ye2022rope3d}.
As the roadside camera is fixed, 
recent methods exploit this fact to approach this problem \cite{jinrang2024monouni, shi2023cobev, yang2023BEVHeight, yang2023bevheight++, wang2024bevspread}.
For example,
BEVHeight \cite{yang2023BEVHeight} trains a model to regress towards object height, which is then converted to depth with camera extrinsics.
BEVSpread \cite{wang2024bevspread} further considers the position approximation error in the voxel pooling process based on BEVHeight.
Differently, our work exploits a 2D detector (Fig.~\ref{fig:overview}b) to produce 2D detections, using which to facilitate 3D detection (Fig.~\ref{fig:overview}a).
Further, we derive a scene prior by exploiting the fixed camera pose, and incorporate the scene prior to enhance  3D detection.
Our methods can serve as a plug-in module to improve existing methods (Table~\ref{tab:benchmarking-results(0.5,0.25,0.25)}).

\textbf{2D object detection} aims to detect objects of predefined categories in a 2D image~\cite{felzenszwalb2009object, lin2014coco}. 
Prevailing approaches train deep neural networks, which typically consist of a backbone to encode input images, and a detector-head for box regression and classification \cite{NIPS2015_fasterRCNN, redmon2016you, liu2016ssd, wang2022yolov7}.
Recently, using the transformer architecture significantly improves 2D detection \cite{detr2020carion, zhu2020deformable, zhang2022dino}.
In this work, we explore how to leverage a 2D detector~\cite{zhang2022dino, wang2022yolov7}  to train better 3D detectors on roadside scenes.
It is important to note that any 2D detectors can be used in our framework Pro3D. 

\begin{figure}[t]
\centering
\includegraphics[trim=0cm 0 0cm 0cm, clip, width=1\linewidth]
{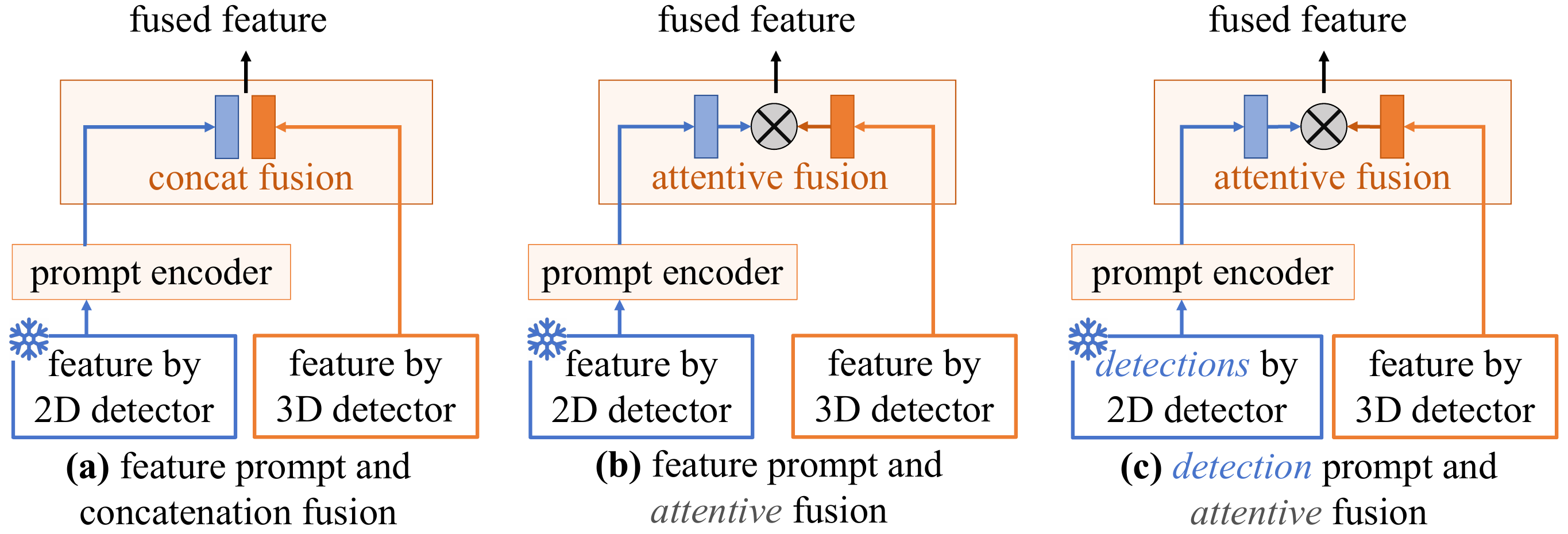}
\vspace{-6.2mm}
\caption{\small
We study three methods for \emph{encoding} and \emph{fusing} 2D detection prompts. 
Design {\bf (a)} concatenates feature maps extracted by the 2D detector and the 3D detector's backbone.
Design {\bf (b)} extracts a feature vector based on a 2D detection's coordinates, encodes it through a {\em prompt encoder}, and attentively fuses it with the feature map extracted by the 3D detector's backbone. 
Design {\bf (c)} encodes a 2D detection, a 5-dim vector (coordinates $x$ and $y$, object width $w$ and height $h$, and the predicted class label) as the prompt, and attentively fuses the encoded detection with the feature map of the 3D detector's backbone.
Somewhat surprisingly, the third performs the best (Table~\ref{tab:results-of-different-prompt-information})!
}
\label{fig:condition by 2D detection}
\vspace{-4mm}
\end{figure}

\textbf{Prompting},
enabled by large language models (LLMs), 
is first recognized in Natural Language Processing (NLP) \cite{brown2020language, sanh2021multitask, wei2022emergent}.
Given an LLM, using a text prompt can instruct the LLM to produce desired text outputs.
Inspired by this, the computer vision community endeavors to train large visual/multimodal models that can take text prompts to generate, restore and edit images \cite{ramesh2021zero, liu2023explicit, nichol2021glide, Sun2024alpha}.
Some works use images as prompts to accomplish vision tasks such as segmentation and inpainting \cite{bar2022visual, wang2023images}.
Beyond visual and text prompts, 
some recent works use more diverse information \cite{wang2025attention} as prompts to guide models to produce desired output.
For example, the Segment Anything Model (SAM) accepts bounding boxes, segmentation masks, and  points as prompts towards interactive segmentation
\cite{kirillov2023segment}.
In this work, we propose the promptable 3D detector design (Pro3D), which trains a 3D detector that accepts 2D detections as prompts.
The instantiation of the prompt can be predicted box coordinates and class labels.

\section{Promptable 3D Detector}
\label{sec:methods}

We start with the problem definition of roadside monocular 3D object detection.
Then, we introduce our \emph{Promptable 3D detector} (Pro3D), including
its crucial modules and a simple technique to obtain scene prior to improve detection performance.
Lastly, we present insightful discussions.

{\bf Problem definition.}
Roadside monocular 3D detection requires detecting objects in a 2D RGB image and predicting their 3D information  $(x, y, z, w, h, l, yaw)$, where $(x, y, z)$ is the object’s center location, $(w, h, l)$ denotes its width, height and length of the cuboid capturing the object, and $yaw$ denotes its orientation on the BEV plane.

\subsection{The Proposed Framework: Pro3D}
\label{ssec:Pro3D}

Pro3D has two core modules to exploit 2D detections to facilitate 3D detection (Fig.~\ref{fig:overview}a): a \emph{prompt encoder} and a \emph{fusion module}.
The former encodes 2D detections as new features to prompt the 3D detector.
The latter attentively fuses the encoded prompts with the 3D detector backbone for 3D detection.
The 3D detector head makes 3D predictions based on the fused features.
We elaborate on the two modules below,
followed by scene prior generation.

{\bf Prompt encoder} treats the output of a 2D detector (either features or boxes) as prompt, and transforms it as a new feature which is exploited for 3D detections. 
The prompt encoder can be implemented as either a set of convolution layers or a multi-layer perceptron (MLP), depending on what are used as prompts and how to fuse (explained later in the \textit{fusion module}).
In our work,
we compare three different prompts extracted from the 2D detector:
(1) global feature maps represented as tensors,
(2) local feature vectors at the spatial position specified by the predicted coordinates of 2D detections,
and (3) the 2D detection consisting of box coordinates $x$ and $y$, width, height, and predicted class label.
If feature maps are used as prompts, the prompt encoder can be a convolutional network that transforms the feature maps with appropriate dimensions (Fig. \ref{fig:condition by 2D detection}a), ensuring that the transformed feature maps can be fused with those of the 3D detector's backbone.
If a 2D detection is used as prompt, 
we represent the 2D box with a normalized positional encoding of its top-left and bottom-right corners, and its predicted class label as a class ID.
We learn an MLP to transform them into a feature vector, using which to attentively fuse the feature map of the 3D detector backbone (details in supplement). 
Somewhat surprisingly, the third prompt performs the best (Table~\ref{tab:results-of-different-prompt-information})!
We explain next.

{\em Remark 1: Using 2D detections as prompts yields better 3D detection performance than using features.}
To explain this, we first note different features extracted from a 2D detector: (1) global feature maps,
and (2) local features at a specific position of the feature maps.
Global feature maps contain all information from the input image, beneficial to object localization but lacking salient signals for semantic prediction.
In contrast, local features can be informative for semantic prediction but lack spatial information for precise localization.
Different from these features, the simplistic 2D detections (consisting of box coordinates, object sizes, and predicted labels) can well localize target objects in the input image and explicitly deliver their well-predicted labels.
By incorporating the 2D detections, a 3D detector can be trained to lift them to the 3D space towards 3D detections. This simplifies the task of roadside monocular 3D detection and facilitates 3D detector training.

\begin{figure}[t]
\centering
\includegraphics[trim=0cm 0 0cm 0cm, clip, width=0.95\linewidth]{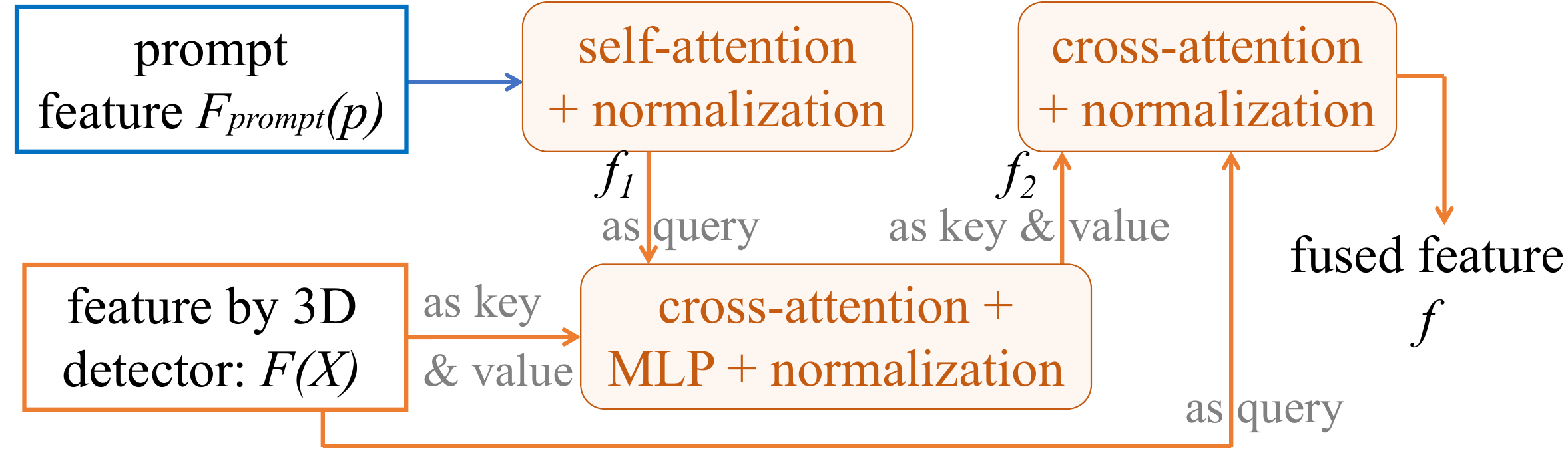}
\vspace{-2.5mm}
\caption{\small
Diagram of our proposed fusion module, which attentively fuse the prompt $p$ (represented by its feature $F_{prompt}(p)$) and feature map $F(X)$ (from the 3D detector's backbone) and ouptut the fused feature $f$.
}
\vspace{-4mm}
\label{fig:fusion_module}
\end{figure}

{\em Remark 2: Different from existing methods, our Pro3D explicitly leverages 2D detections.}
Leveraging 2D box \emph{annotations} to facilitate 3D detector training is a \emph{de facto} practice.
A  similar work to ours is MV2D~\cite{wang2023object},
which trains a 3D detector from scratch for joint 2D and 3D detection.
Specifically, MV2D fuses RoI features extracted from the 2D detector with those of the 3D detector's backbone. This is similar in spirit to our explored baseline method (Fig. \ref{fig:condition by 2D detection}b). But our final version of Pro3D explicitly leverages 2D detections instead of their features.
Extensive experiments show that Pro3D significantly outperforms MV2D (Table~\ref{tab:benchmarking-results(0.5,0.25,0.25)}). Importantly, owing to the modular design, Pro3D can readily exploit any 2D detectors  -- using more advanced ones yields better 3D detection performance (Fig.~\ref{fig:YOLO-v7 and more 2D annotations}).


{\bf Fusion module}
fuses the encoded prompt and the feature maps of the 3D detector backbone (Fig.~\ref{fig:condition by 2D detection}).
If the prompt is represented by a feature map extracted from the 2D detector, a straightforward fusion method is to concatenate these feature maps (Fig.~\ref{fig:condition by 2D detection}a), as widely done in the literature \cite{long2015fully, pinheiro2016learning}.
Our final method fuses feature vectors which encode 2D detections as prompts.
In this work, we implement a  Transformer network \cite{vaswani2017attention, kirillov2023segment} to attentively fuse the feature vectors with feature maps of the 3D detector backbone (Fig.~\ref{fig:fusion_module}).
Concretely, for an input image $X$ which goes through the 3D detector backbone $F$, 
its feature map  is denoted as $F(X)$.
The prompt encoder $F_{prompt}$ transforms a  prompt $p$ into a vector $F_{prompt}(p)$.
The fusion model returns the final fused feature 
$f= \text{Norm}(\text{CrossAttn}(\text{query}=F(X), \text{key}=f_2, \text{value}=f_2))$,
where 
\begin{equation}\small
\begin{aligned}
f_2 = \text{Norm}(&\text{MLP}(\text{CrossAttn}( \text{query}=f_1, \\
&\text{key}=F(X), \text{value}=F(X)))) \\
f_1 = \text{Norm}(&\text{SelfAttn}(F_{\text{prompt}}(p))) \\
\end{aligned}
\nonumber
\end{equation}
Experiments demonstrate that such attentive fusion outperforms concatenation (Table~\ref{tab:results-of-different-prompt-information}).

\begin{figure}[t]
\centering
\includegraphics[trim=0cm 0 0cm 0cm, clip, width=1\linewidth]{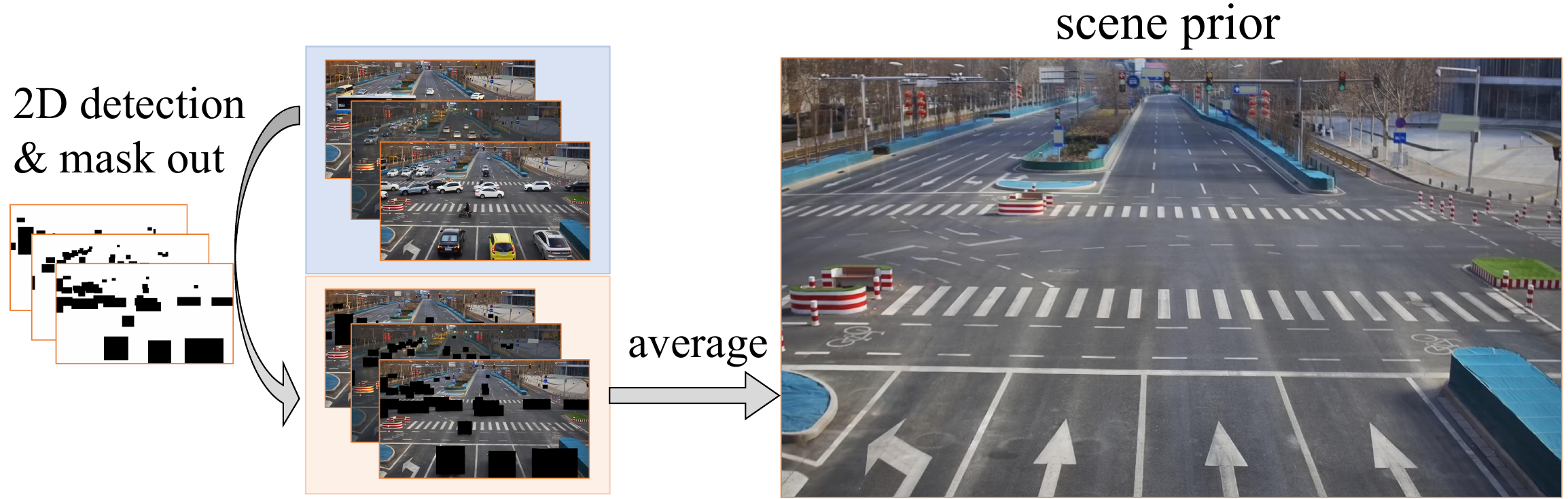}
\vspace{-7mm}
\caption{\small
{\bf Scene prior generation.}
We mask out objects of interest in training frames belonging to a specific scene, and average the remaining pixels across frames towards an ``empty'' background, which is our scene prior.
A 3D detector that incorporates this scene prior (Fig.~\ref{fig:overview}a) achieves remarkable improvements in roadside monocular 3D detection (Table \ref{tab:background}).
}
\label{fig:background}
\vspace{-4mm}
\end{figure}

{\bf Scene prior.}
Hung on a roadside perception infrastructure,
the camera is fixed \cite{zhao2021camera}. Hence, the scene geometry in the captured video frames can be assumed to be stable.
This inspires us to derive a scene prior from these frames.
To generate a scene prior,
we mask out objects (by running a 2D detector on the training frames belonging to a specific scene),
and then average the remaining pixels across frames to obtain an ``empty'' background image for this scene, which is the scene prior (Fig.~\ref{fig:background}).
In this work, we simply concatenate the scene prior with the RGB frame as a six-channel input to the 3D detector.
We study the affects of using different scene priors generated in daytime and nighttime, respectively;
results in Table~\ref{tab:results-of-priors} show that using either remarkably enhances performance.

\subsection{Discussions on 2D Detector Training} 

As Pro3D explicitly exploits a 2D detector (Fig.~\ref{fig:overview}a),
in principle, it can use any off-the-shelf powerful 2D detectors  \cite{Ma2023LongTailed3D}.
Below we make three remarks on this matter.

{\bf Techniques of 2D detection are well established.} 
2D detection is more sufficiently explored than monocular 3D detection, as the former has been a fundamental problem in computer vision whereas the latter is relatively new.
There are many excellent 2D detectors, such as FasterRCNN \cite{NIPS2015_fasterRCNN},
YOLO \cite{bochkovskiy2020yolov4, wang2022yolov7}, 
DETR \cite{detr2020carion} and DINO \cite{zhang2022dino}.
In this work,
we use the well-known transformer-based detector DINO \cite{zhang2022dino}.
We also test other 2D detectors in experiments.

{\bf More 2D detection datasets are publicly available,} 
while the relatively new problem of monocular 3D detection has less datasets which are also small in scale.
For example, the 2D detection dataset COCO \cite{lin2014coco}  (published in 2014) annotates 330K images with 2D boxes for 80 classes,
whereas the 3D detection dataset nuScenes \cite{caesar2020nuscenes} (published in 2020) ``only'' annotates 144K images with 3D cuboids for 23 classes. 
The vast amounts of 2D annotated data allows pretraining stronger 2D detectors,
using which Pro3D yields more remarkable 3D detection performance gains  (Fig.~\ref{fig:YOLO-v7 and more 2D annotations}).


\textbf{Joint training \emph{vs.} stage-wise training.}
For 3D detection, most methods jointly train the 3D detector and a 2D detector by incorporating 2D annotations in an auxiliary loss \cite{li2023BEVDepth, liu2022learning, park2021pseudo, simonelli2019disentangling, wang2022mv, wang2021fcos3d, zhang2021objects, guo2021liga, yang2023bevformer, wang2023object,luo2023latr}.
Our Pro3D is a stage-wise training method, which turns out to outperform joint training (Table~\ref{tab:ablation-joint-training}).
We conjecture this is due to two reasons.
First, 2D detector training is better explored with mature techniques than 3D detector training, so stage-wise training makes use of such techniques of 2D detector training.
Second, joint training must strike a balance between various loss functions such as 3D cuboid regression, rotation regression and so on; yet, optimizing over all losses inevitably sacrifices 2D detection which is actually much easier than 3D detection.
It is worth noting that stage-wise training allows easy leverage of  external 2D datasets for pretraining a stronger 2D detector, exploiting which helps our Pro3D achieve even better performance  (Fig.~\ref{fig:YOLO-v7 and more 2D annotations}).

\begin{figure*}[t]
\centering
\includegraphics[trim=0cm 0 0cm 0cm, clip, width=0.99\linewidth, height=3.5cm]{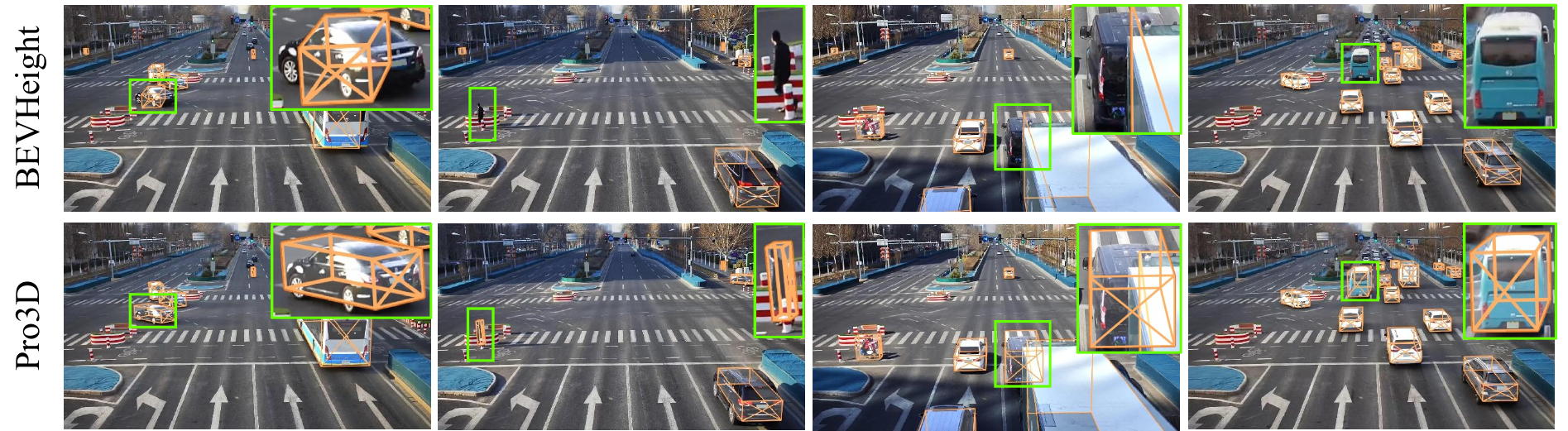}
\vspace{-3mm}
\caption{\small
Visual comparison between the state-of-the-art method BEVHeight \cite{yang2023BEVHeight} and our Pro3D.
Results show that Pro3D can (1) make better orientation predictions than BEVHeight (column-1),
and (2) detect objects (missed by BEVHeight) that are too small in size (column-2), heavily occluded (column-3), and in the far field (column-4).
}
\vspace{-3mm}
\label{fig:visualization of Pro3D}
\end{figure*}

{
\setlength{\tabcolsep}{0.710mm}
\begin{table}[t]
\small
\centering
\caption{\small
{\bf Benchmarking on DAIR-V2X-I.}
Following \cite{yang2023BEVHeight}, we report AP metrics at IoU=0.5, 0.25, 0.25 in BEV on Vehicle, Cyclist, and Pedestrian, respectively. We mark the modalties (M) used in each method: C for RGB camera, L for LiDAR.
As our Pro3D can take in any 3D detectors, we build it with BEVDepth, BEVHeight, and BEVSpread, respectively.
Results show that  Pro3D significantly improves these methods, e.g., it helps BEVHeight achieve $>$10 AP improvements on Pedestrian and Cyclist! 
}
\vspace{-3mm}
\scalebox{0.70}{
\begin{tabular}{lc ccc ccc ccc}
\toprule
\multirow{2}{*}{Method}            & 
\multirow{2}{*}{M}            & 
\multicolumn{3}{c}{Vehicle} &\multicolumn{3}{c}{Cyclist}
&\multicolumn{3}{c}{Pedestrian} \\
\cmidrule(r){3-5} \cmidrule(r){6-8} \cmidrule(r){9-11} 
& & {\tt Easy}   & {\tt Mid}   & {\tt Hard}   & {\tt Easy}   & {\tt Mid}    & {\tt Hard}
      & {\tt Easy}   & {\tt Mid}    & {\tt Hard}   \\
\midrule
PointPillars \cite{lang2019pointpillars}  &  L   &  63.1   &   54.0       &   54.0    &   38.5    &  22.6      &   22.5       &   38.5    &   37.2  &  37.3   \\
SECOND \cite{yan2018second}  &  L  &   71.5       &   54.0    &   54.0    &  54.7      &   31.1       &   31.2    &   55.2  &  52.5  &  52.5   \\
MVXNet \cite{sindagi2019mvxnet}  &  L+C   &   71.0       &   53.7    &   53.8    &  54.1      &   30.8       &   31.1    &   55.8  &  54.5  &  54.4   \\
\midrule

proj. w/ class prior & C & 33.2 & 24.6 & 24.5 & 15.2 & 14.9 & 14.8 & 13.5 & 13.3 & 13.4 \\
proj. w/ learning  & C & 39.2 & 29.9 & 29.7 & 18.6 & 18.3 & 18.3 & 13.9 & 13.8 & 13.9 \\

\multirow{1}{*}{FCOS3D \cite{wang2021fcos3d}}  &  C &  \multirow{1}{*}{54.2}   &   \multirow{1}{*}{49.9}       &   \multirow{1}{*}{50.1}    &   \multirow{1}{*}{25.6}    &  \multirow{1}{*}{24.3}      &   \multirow{1}{*}{24.4}       &   \multirow{1}{*}{19.2}    &   \multirow{1}{*}{19.0}   & \multirow{1}{*}{19.0}   \\

FCOS3D+ & C & 59.3 & 55.2 & 55.7 & 35.2 & 34.8 & 34.9 & 27.9 & 27.5 & 27.6 \\

\multirow{1}{*}{M3D-RPN \cite{brazil2019m3d}}  & C &  \multirow{1}{*}{49.6}   &   \multirow{1}{*}{40.3}       &   \multirow{1}{*}{40.2}    &   \multirow{1}{*}{22.5}    &  \multirow{1}{*}{20.0}      &   \multirow{1}{*}{20.6}       &   \multirow{1}{*}{12.1}    &   \multirow{1}{*}{12.3}   & \multirow{1}{*}{12.5}   \\
\multirow{1}{*}{Kinematic3D \cite{brazil2020kinematic}} & C &  \multirow{1}{*}{50.6}   &   \multirow{1}{*}{45.7}       &   \multirow{1}{*}{45.6}    &   \multirow{1}{*}{23.6}    &  \multirow{1}{*}{22.1}      &   \multirow{1}{*}{22.3}       &   \multirow{1}{*}{13.5}    &   \multirow{1}{*}{13.6}   & \multirow{1}{*}{13.1}   \\

\multirow{1}{*}{MonoDLE \cite{ma2021delving}}  &  C & \multirow{1}{*}{52.3}   &   \multirow{1}{*}{48.2}       &   \multirow{1}{*}{48.1}    &   \multirow{1}{*}{24.1}    &  \multirow{1}{*}{23.5}      &   \multirow{1}{*}{23.6}       &   \multirow{1}{*}{14.6}    &   \multirow{1}{*}{14.4}   & \multirow{1}{*}{14.5}   \\
\multirow{1}{*}{MonoFlex \cite{zhang2021objects}}  &  C &  \multirow{1}{*}{60.0}   &   \multirow{1}{*}{51.1}       &   \multirow{1}{*}{51.3}    &   \multirow{1}{*}{26.1}    &  \multirow{1}{*}{26.0}      &   \multirow{1}{*}{26.3}       &   \multirow{1}{*}{18.2}    &   \multirow{1}{*}{18.3}   & \multirow{1}{*}{18.5}   \\

\multirow{1}{*}{CaDDN \cite{reading2021categorical}}  & C &  \multirow{1}{*}{51.2}   &   \multirow{1}{*}{46.6}       &   \multirow{1}{*}{46.7}    &   \multirow{1}{*}{23.9}    &  \multirow{1}{*}{22.8}      &   \multirow{1}{*}{22.6}       &   \multirow{1}{*}{14.0}    &   \multirow{1}{*}{13.9}   & \multirow{1}{*}{13.9}   \\

\multirow{1}{*}{MonoNeRD \cite{xu2023mononerd}} 
 & C  &  \multirow{1}{*}{65.2}   &   \multirow{1}{*}{57.3}       &   \multirow{1}{*}{57.0}    &   \multirow{1}{*}{40.0}    &  \multirow{1}{*}{40.2}      &   \multirow{1}{*}{40.5}       &   \multirow{1}{*}{30.1}    &   \multirow{1}{*}{29.6}   & \multirow{1}{*}{30.2}   \\

\midrule

ImvoxelNet \cite{rukhovich2022imvoxelnet}  & C &   44.8       &   37.6    &   37.6    &  21.1      &   13.6       &   13.2    &   6.8  &  6.7 & 6.7 \\
BEVFormer \cite{li2022bevformer}  & C &   61.4       &   50.7    &   50.7    &  22.2      &   22.1       &   22.1    &   16.9  &  15.8  &  16.0   \\
BEVFormer v2 \cite{yang2023bevformer}  & C &  62.9       &   52.3    &   52.3    &  25.3      &   25.2       &  25.2    &   18.0  &  17.1  &  17.2   \\
SparseBEV  \cite{liu2023sparsebev}  & C &   65.6       &   55.1    &   52.6    & 30.0      &  30.3      &   30.0    &   18.4  &  17.9  &  17.5   \\
MV2D \cite{wang2023object}  & C &   65.7       &   57.1    &   57.0    &  43.8     &   42.9       &   43.1    &   29.0  &  28.3  &  29.1   \\
MonoGAE  \cite{yang2024monogae}  & C &   84.6       &   \textbf{75.9 }   &   \textbf{74.2}    &  44.0      &   47.6      &   46.8    &   25.7  &  24.3 &  24.4   \\
BEVHeight++  \cite{yang2023bevheight++}  & C &   79.3       &   68.6    &   68.7     &   60.8  &  60.5  &  61.0  & 42.9      &  40.9      &   41.1    \\
\midrule

BEVDepth \cite{li2023BEVDepth}  & C &   75.5       &   63.6    &   63.7       &   55.7  &  55.5  &  55.3  &  35.0      &   33.4       &   33.3 \\
{\bf Pro3D} w/ BEVDepth  & C &   79.6       &   71.8    &   71.7    &  65.8      &   65.8       &   65.3    &   46.0  &  44.7  &  44.9   \\
\quad \emph{performance gain} & C &  \multirow{1}{*}{\color{darkgreen}+4.1}   &   \multirow{1}{*}{\color{darkgreen}+8.2}       &   \multirow{1}{*}{\color{darkgreen}+8.0}    &   \multirow{1}{*}{\color{darkgreen}+10.1}    &  \multirow{1}{*}{\color{darkgreen}+10.3}      &   \multirow{1}{*}{\color{darkgreen}+10.0}       &   \multirow{1}{*}{\color{darkgreen}+11.0}    &   \multirow{1}{*}{\color{darkgreen}+11.3}   & \multirow{1}{*}{\color{darkgreen}+11.6} \\
\midrule
BEVHeight \cite{yang2023BEVHeight}  & C &   77.8       &   65.8    &   65.9    &  60.2      &   60.1      &   60.5    &   41.2  &  39.3 &  39.5   \\
{\bf Pro3D} w/ BEVHeight  & C &   83.7       &   71.8    &   72.0    &  71.3      &   71.2       &   71.4    &   53.9  &  52.9  &  52.6   \\
\quad \emph{performance gain} & C &  \multirow{1}{*}{\color{darkgreen}+5.9}   &   \multirow{1}{*}{\color{darkgreen}+6.0}       &   \multirow{1}{*}{\color{darkgreen}+6.1}    &   \multirow{1}{*}{\color{darkgreen}+11.1}    &  \multirow{1}{*}{\color{darkgreen}+11.1}      &   \multirow{1}{*}{\color{darkgreen}+10.9}       &   \multirow{1}{*}{\color{darkgreen}+12.7}    &   \multirow{1}{*}{\color{darkgreen}+13.6}   & \multirow{1}{*}{\color{darkgreen}+13.1} \\
\midrule
BEVSpread \cite{wang2024bevspread}  & C &   79.1       &   66.8    &   66.9    &  62.6      &   63.5      &   63.8    &   46.5  &  44.5 &  44.7   \\
{\bf Pro3D} w/ BEVSpread   & C &   \textbf{85.2}       &   73.1    &   73.6    &  \textbf{73.1}      &   \textbf{73.0}       &   \textbf{73.1}    &   \textbf{55.1}  &  \textbf{54.3}  &  \textbf{54.1}   \\
\quad  \emph{performance gain} & C &  \multirow{1}{*}{\color{darkgreen}+6.1}   &   \multirow{1}{*}{\color{darkgreen}+6.3}       &   \multirow{1}{*}{\color{darkgreen}+6.7}    &   \multirow{1}{*}{\color{darkgreen}+10.5}    &  \multirow{1}{*}{\color{darkgreen}+9.5}      &   \multirow{1}{*}{\color{darkgreen}+9.3}       &   \multirow{1}{*}{\color{darkgreen}+8.6}    &   \multirow{1}{*}{\color{darkgreen}+9.8}   & \multirow{1}{*}{\color{darkgreen}+9.4} \\

\bottomrule
\end{tabular}
}
\vspace{-5mm}
\label{tab:benchmarking-results(0.5,0.25,0.25)}
\end{table}
}

{
\setlength{\tabcolsep}{1.5mm}
\begin{table}[t]
\small
\centering
\caption{\small
{\bf Benchmarking results on the Rope3D dataset.}
Following \cite{ye2022rope3d},
we use metrics AP and {\tt Rope} at IoU=0.5 and 0.7.
Built with detectors BEVHeight and BEVSpread, Pro3D achieves significant improvements.
Notably, it improves the original BEVHeight and BEVSpread by $>$26 AP on Big Vehicle! 
Conclusions in Table~\ref{tab:benchmarking-results(0.5,0.25,0.25)} hold on this dataset.
}
\vspace{-3mm}
\scalebox{0.73}{
\begin{tabular}{l cccc cccc}
\toprule
\multirow{3}{*}{Method}            &  
\multicolumn{4}{c}{IoU = 0.5} &\multicolumn{4}{c}{IoU = 0.7} \\
\cmidrule(l){2-5} \cmidrule(l){6-9} 
&  \multicolumn{2}{c}{Car} & \multicolumn{2}{c}{Big Vehicle} & \multicolumn{2}{c}{Car} & \multicolumn{2}{c}{Big Vehicle} \\
\cmidrule(l){2-3} \cmidrule(l){4-5} 
\cmidrule(l){6-7} \cmidrule(l){8-9} 
& {\tt AP}   & {\tt Rope}   & {\tt AP}   & {\tt Rope}   & {\tt AP}   & {\tt Rope}   & {\tt AP}   & {\tt Rope} \\
\midrule

M3D-RPN \cite{brazil2019m3d}    &  54.2   &   62.7       &   33.1    &   44.9 &  16.8   &   32.9       &   6.9    &   24.2      \\
Kinematic3D \cite{brazil2020kinematic}  &  50.6   &   58.9       &   37.6    &   48.1  &  17.7   &   32.9       &   6.1    &   22.9      \\
MonoDLE \cite{ma2021delving}   &  51.7   &   60.4       &   40.3    &   50.1 &  17.7   &   32.9       &   6.1    &   22.9      \\
MonoFlex \cite{zhang2021objects}   &  60.3   &   66.9       &   37.3    &   48.0  &  33.8   &   36.1       &   10.1    &   26.2      \\
BEVFormer \cite{li2022bevformer}    &  50.6   &   58.8       &   34.6    &   45.2 &  24.6   &   38.7       &   10.1    &   25.6      \\
SparseBEV  \cite{liu2023sparsebev}     &  54.7   &   62.3       &   39.8    &   48.4 &  29.9   &   41.2       &   14.6  &   30.0    \\
BEVDepth \cite{li2023BEVDepth}  &  69.6   &   74.7       &   45.0    &   54.6    &  42.6   &   53.1       &   21.5    &   35.8      \\
BEVHeight++ \cite{yang2023bevheight++}  &  76.1   &   80.9       &   50.1   &   59.9    &  47.0   &   57.7       &   24.4    &   39.6      \\
\midrule

BEVHeight \cite{yang2023BEVHeight}  &  74.6   &   78.7       &   48.9    &   57.7   &  45.7   &   55.6       &   23.1    &   37.0      \\
\textbf{Pro3D} w/ BEVHeight  &    84.0       &    86.2    &   75.1    &  79.2 &    51.0       &    59.2    &   32.1    &  44.3     \\
\quad \emph{performance gain} &  \multirow{1}{*}{\color{darkgreen}+9.4}   &   \multirow{1}{*}{\color{darkgreen}+7.5}       &   \multirow{1}{*}{\color{darkgreen}+26.2}    &   \multirow{1}{*}{\color{darkgreen}+21.5}    &  \multirow{1}{*}{\color{darkgreen}+5.3}      &   \multirow{1}{*}{\color{darkgreen}+3.6}       &   \multirow{1}{*}{\color{darkgreen}+11.0}    &   \multirow{1}{*}{\color{darkgreen}+7.3}  \\
\midrule
BEVSpread \cite{wang2024bevspread}  &  76.9   &   80.8       &   51.2    &   60.3   &  47.1   &   57.6       &   25.6    &   39.3      \\
\textbf{Pro3D} w/ BEVSpread   &    \textbf{86.3}       &    \textbf{87.6}    &   \textbf{77.6}    &  \textbf{81.0} &    \textbf{53.1}       &    \textbf{60.6}    &   \textbf{33.9}    &  \textbf{45.6}     \\
\quad \emph{performance gain} &  \multirow{1}{*}{\color{darkgreen}+9.4}   &   \multirow{1}{*}{\color{darkgreen}+6.8}       &   \multirow{1}{*}{\color{darkgreen}+26.4}    &   \multirow{1}{*}{\color{darkgreen}+20.7}    &  \multirow{1}{*}{\color{darkgreen}+6.0}      &   \multirow{1}{*}{\color{darkgreen}+3.0}       &   \multirow{1}{*}{\color{darkgreen}+8.3}    &   \multirow{1}{*}{\color{darkgreen}+6.3}  \\
\bottomrule
\end{tabular}
}
\vspace{-4mm}
\label{tab:benchmarking-results of Rope3d(0.7,0.5,0.5)}
\end{table}
}

\section{Experiments}

We validate Pro3D through extensive experiments.
We first describe experiment settings including datasets, metrics and implementation details.
We then compare Pro3D against existing works.
We lastly conduct comprehensive ablation study.
The supplement provide more analyses.

\subsection{Settings}
\label{ssec:setting}

{\bf Datasets.}
We use two established datasets:
\begin{itemize}[topsep=-2pt, partopsep=3pt, leftmargin=0.1in]
    \item \emph{DAIR-V2X-I} \cite{yu2022dair} is the first dataset of roadside monocular 3D detection. 
    It provides 5K and 2K images for training and validation, respectively.
    Following \cite{yang2023BEVHeight},
    we report results on the validation set, with breakdown analysis w.r.t three superclasses (vehicle, cyclist, and pedestrian), at three difficulty levels ({\tt Easy}, {\tt Medium}, {\tt Hard}).
        
    \item \emph{Rope3D} \cite{ye2022rope3d} is another large-scale dataset that contains 50K images and annotations of $>$1.5M objects in diverse scenes. Following \cite{yang2023BEVHeight}, we use its train-set (70\%) to train models and the validation set (30\%) for benchmarking. 
    \vspace{2mm}
\end{itemize}

{\bf Metrics.} 
We introduce benchmarking metrics below.
\begin{itemize}[topsep=-3pt, partopsep=3pt, leftmargin=0.1in]
\item  
    AP$_{\text{IoU}=t}$ is the average precision (AP) at IoU threshold $t$ on the BEV plane.
    Following \cite{yang2023BEVHeight}, we set $t=0.25$ for ``cyclist'' and ``pedestrian'', and  $t=0.5$ for ``vehicle''.
   
\item 
    Mean AP (mAP)  \cite{lin2014coco} measures 2D detection performance on the image plane (Fig.~\ref{fig:YOLO-v7 and more 2D annotations}).
    It first computes per-class average precision  (AP) over IoU thresholds from 0.5 to 0.95 with stepsize 0.05, and then averages per-class APs.
\item 
    On the Rope3D benchmark dataset \cite{ye2022rope3d},
    we report numbers w.r.t its official metrics of AP$_{\text{IoU}=t}$ in  BEV \cite{simonelli2019disentangling}
    and Rope$_{score}$.
    The latter is a sophisticated metric that jointly considers detection errors of orientation, area, and BEV box coordinates compared to the ground-truth.
\vspace{1.5mm}
\end{itemize}

{\bf Compared methods.}
Pro3D can take in any 2D detectors and 3D detectors.
To show the benefits of Pro3D,
we build it with representative roadside 3D detectors,
including 
BEVDepth~\cite{li2023BEVDepth},
BEVHeight~\cite{yang2023BEVHeight}, and BEVSpread~\cite{wang2024bevspread}.
For benchmarking, we compare against more methods which are either published recently or representative in the literature 3D detection.
We refer the reader to Table~\ref{tab:benchmarking-results(0.5,0.25,0.25)} and \ref{tab:benchmarking-results of Rope3d(0.7,0.5,0.5)} for the compared methods.

{
\setlength{\tabcolsep}{0.95mm}
\begin{table}[t]
\small
\centering
\caption{\small
{\bf Incorporating scene prior remarkably improves roadside 3D detection.}
Interestingly, small objects of Cyclist and Pedestrian get more improvements than Vehicle.
}
\vspace{-3mm}
\scalebox{0.74}{
\begin{tabular}{l ccc ccc ccc}
\toprule
\multirow{2}{*}{Method}            & 
\multicolumn{3}{c}{Vehicle} &\multicolumn{3}{c}{Cyclist}
&\multicolumn{3}{c}{Pedestrian} \\
\cmidrule(r){2-4} \cmidrule(r){5-7} \cmidrule(r){8-10} 
& {\tt Easy}   & {\tt Mid}   & {\tt Hard}   & {\tt Easy}   & {\tt Mid}    & {\tt Hard}
      & {\tt Easy}   & {\tt Mid}    & {\tt Hard}   \\
\midrule
\rowcolor{lightgrey} BEVHeight~\cite{yang2023BEVHeight}  &  \multirow{1}{*}{77.8}   &   \multirow{1}{*}{65.8}       &   \multirow{1}{*}{65.9}    &   \multirow{1}{*}{60.2}    &  \multirow{1}{*}{60.1}      &   \multirow{1}{*}{60.5}       &   \multirow{1}{*}{41.2}    &   \multirow{1}{*}{39.3}   & \multirow{1}{*}{39.5}   \\

\quad+ scene prior &  \multirow{1}{*}{79.2}   &   \multirow{1}{*}{67.7}       &   \multirow{1}{*}{67.8}  &   \multirow{1}{*}{62.5}    &   \multirow{1}{*}{62.3}   & \multirow{1}{*}{62.7}    &   \multirow{1}{*}{44.3}    &  \multirow{1}{*}{42.5}      &   \multirow{1}{*}{42.7} \\

\emph{performance gain} &  \multirow{1}{*}{\color{darkgreen}+1.4}   &   \multirow{1}{*}{\color{darkgreen}+1.9}       &   \multirow{1}{*}{\color{darkgreen}+1.9}    &   \multirow{1}{*}{\color{darkgreen}+2.3}    &  \multirow{1}{*}{\color{darkgreen}+2.2}      &   \multirow{1}{*}{\color{darkgreen}+2.2}       &   \multirow{1}{*}{\color{darkgreen}+3.1}    &   \multirow{1}{*}{\color{darkgreen}+3.2}   & \multirow{1}{*}{\color{darkgreen}+3.2}   \\

\midrule

\rowcolor{lightgrey} BEVSpread~\cite{wang2024bevspread}  &  \multirow{1}{*}{79.1}   &   \multirow{1}{*}{66.8}       &   \multirow{1}{*}{66.9}    &   \multirow{1}{*}{62.6}    &  \multirow{1}{*}{63.5}      &   \multirow{1}{*}{63.8}       &   \multirow{1}{*}{46.5}    &   \multirow{1}{*}{44.5}   & \multirow{1}{*}{44.7}   \\

\quad+ scene prior &  \multirow{1}{*}{80.1}   &   \multirow{1}{*}{68.5}       &   \multirow{1}{*}{68.6}  &   \multirow{1}{*}{64.5}    &   \multirow{1}{*}{65.4}   & \multirow{1}{*}{65.6}    &   \multirow{1}{*}{49.6}    &  \multirow{1}{*}{47.5}      &   \multirow{1}{*}{47.7} \\

\emph{performance gain} &  \multirow{1}{*}{\color{darkgreen}+1.0}   &   \multirow{1}{*}{\color{darkgreen}+1.7}       &   \multirow{1}{*}{\color{darkgreen}+1.7}    &   \multirow{1}{*}{\color{darkgreen}+1.9}    &  \multirow{1}{*}{\color{darkgreen}+1.9}      &   \multirow{1}{*}{\color{darkgreen}+1.8}       &   \multirow{1}{*}{\color{darkgreen}+3.1}    &   \multirow{1}{*}{\color{darkgreen}+3.0}   & \multirow{1}{*}{\color{darkgreen}+3.0}   \\

\midrule

\rowcolor{lightgrey} Pro3D w/ BEVSpread&  \multirow{1}{*}{84.3}   &   \multirow{1}{*}{71.5}       &   \multirow{1}{*}{72.0}  &   \multirow{1}{*}{71.4}    &   \multirow{1}{*}{71.3}   & \multirow{1}{*}{71.5}    &   \multirow{1}{*}{52.2}    &  \multirow{1}{*}{51.6}      &   \multirow{1}{*}{51.4} \\

\quad+ scene prior &  \multirow{1}{*}{85.2}   &   \multirow{1}{*}{73.1}       &   \multirow{1}{*}{73.6}  &   \multirow{1}{*}{73.1}    &   \multirow{1}{*}{73.0}   & \multirow{1}{*}{73.1}    &   \multirow{1}{*}{55.1}    &  \multirow{1}{*}{54.3}      &   \multirow{1}{*}{54.1} \\

\emph{performance gain} &  \multirow{1}{*}{\color{darkgreen}+0.9}   &   \multirow{1}{*}{\color{darkgreen}+1.6}       &   \multirow{1}{*}{\color{darkgreen}+1.6}    &   \multirow{1}{*}{\color{darkgreen}+1.7}    &  \multirow{1}{*}{\color{darkgreen}+1.7}      &   \multirow{1}{*}{\color{darkgreen}+1.6}       &   \multirow{1}{*}{\color{darkgreen}+2.9}    &   \multirow{1}{*}{\color{darkgreen}+2.7}   & \multirow{1}{*}{\color{darkgreen}+2.7}   \\

\bottomrule
\end{tabular}
}
\vspace{-2mm}
\label{tab:background}
\end{table}
}

{\bf Implementations.}
3D detectors studied in our work mostly follow BEVHeight \cite{yang2023BEVHeight} and BEVSpread \cite{wang2024bevspread}, 
both of which have a Res101-based image encoder \cite{he2016deep},
a Res18-based BEV encoder, 
an FPN for proposal detection  \cite{yan2018second},
and a detector head for the final output of 3D detections.
Note that our Pro3D replaces the Res18-based BEV encoder with a single convolutional layer, removing the FPN as Pro3D uses 2D detection already. This modification helps Pro3D achieve faster training and inference than previous methods (Table~\ref{tab:inference-time} \& \ref{tab:training_time}).
Within the Pro3D framework, we report performance by using BEVDepth, BEVHeight and BEVSpread as the 3D detection module, respectively.
owing to BEVHeight's fast inference speed (detailed in the supplement), we use it in Pro3D for ablation study.
For 2D detector, we train DINO~\cite{zhang2022dino} with the backbone ViT-S/16 as the 2D detector using the default hyperparameter setting.
For all methods, the input image resolution is 864$\times$1536 and the initial voxel resolution is 1024$\times$1024$\times$1. 
We adopt the AdamW optimizer \cite{loshchilov2017decoupled} and learning rate 8e-4.
During training, we adopt random scaling and rotation for data augmentation.
We select the best checkpoint via validation.
We remove detections that have confidence score less than 0.3.


{
\setlength{\tabcolsep}{1.mm}
\begin{table}[t]
\small
\centering
\caption{\small  
{\bf Comparison of different scene priors for our Pro3D.} 
Using scene prior generated by using either all training images or those captured at night enhances the final performance}
\vspace{-3.3mm}
\scalebox{0.71}{
\begin{tabular}{l ccc ccc ccc}
\toprule
\multirow{2}{*}{Method}            & 
\multicolumn{3}{c}{Vehicle} &\multicolumn{3}{c}{Cyclist}
&\multicolumn{3}{c}{Pedestrian} \\
\cmidrule(r){2-4} \cmidrule(r){5-7} \cmidrule(r){8-10}
& {\tt Easy}   & {\tt Mid}   & {\tt Hard}   & {\tt Easy}   & {\tt Mid}    & {\tt Hard}
      & {\tt Easy}   & {\tt Mid}    & {\tt Hard}   \\
\midrule
\multirow{1}{*}{Pro3D w/ BEVSpread~\cite{wang2024bevspread}}  &  \multirow{1}{*}{84.3}   &   \multirow{1}{*}{71.5}       &   \multirow{1}{*}{72.0}    &   \multirow{1}{*}{71.4}    &  \multirow{1}{*}{71.3}      &   \multirow{1}{*}{71.5}       &   \multirow{1}{*}{52.2}    &   \multirow{1}{*}{51.6}   & \multirow{1}{*}{51.4}   \\
\multirow{1}{*}{{\ \ \ + scene priors (All)}}  &  \multirow{1}{*}{{85.2}}   &   \multirow{1}{*}{{73.1}}       &   \multirow{1}{*}{{73.6}}    &   \multirow{1}{*}{{73.1}}    &  \multirow{1}{*}{{73.0}}      &   \multirow{1}{*}{{73.1}}       &   \multirow{1}{*}{{55.1}}    &   \multirow{1}{*}{{54.3}}   & \multirow{1}{*}{{54.1}}   \\
\multirow{1}{*}{\ \ \ + scene priors (Night)}  &  \multirow{1}{*}{85.0}   &   \multirow{1}{*}{72.7}       &   \multirow{1}{*}{73.0}    &   \multirow{1}{*}{72.7}    &  \multirow{1}{*}{72.5}      &   \multirow{1}{*}{72.6}       &   \multirow{1}{*}{53.9}    &   \multirow{1}{*}{53.6}   & \multirow{1}{*}{53.4}   \\

\bottomrule
\end{tabular}
}
\vspace{-2mm}
\label{tab:results-of-priors}
\end{table}
}

\begin{figure}[t]
\centering
\includegraphics[trim=0cm 0 0cm 0cm, clip, width=1\linewidth]
{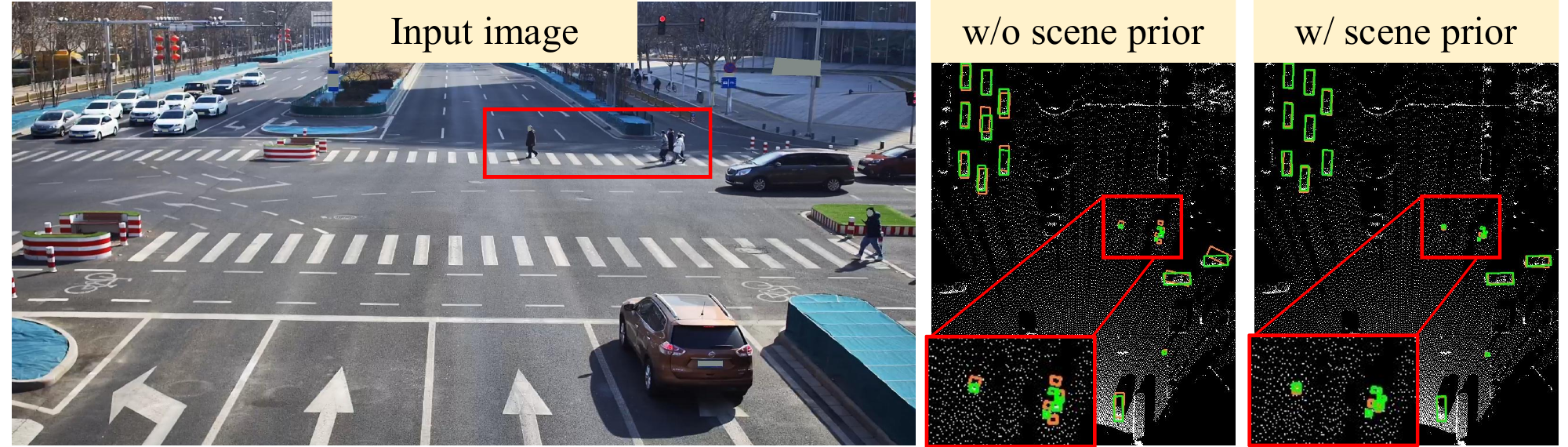}
\vspace{-7mm}
\caption{\small 
Comparison of 3D detections by Pro3D on a test image of the DAIR-V2X-I dataset.
Ground-truth and predictions are green and red, respectively. 
The area in red boxes reveals a clear improvement in pedestrian (i.e., small objects) detection when the scene prior is used. Zoom in to see better.
}
\label{fig:visualcompare_scene_prior}
\vspace{-4mm}
\end{figure}

\subsection{Benchmarking Results}

We study how much improvements our Pro3D brings to existing roadside 3D detectors on DAIR-V2X-I (Table~\ref{tab:benchmarking-results(0.5,0.25,0.25)})
and Rope3D (Table~\ref{tab:benchmarking-results of Rope3d(0.7,0.5,0.5)}).
Fig.~\ref{fig:visualization of Pro3D} visualizes results by Pro3D and BEVHeight (refer to the supplement for more visual comparisons).
The results demonstrate that our Pro3D helps
previous methods BEVDepth \cite{li2023BEVDepth}, BEVHeight \cite{yang2023BEVHeight}, and BEVSpread \cite{wang2024bevspread}
achieve significant performance gains.
Importantly, 
Pro3D boosts 3D detection performance particularly on small objects of ``cyclist'' and ``pedestrian''.

A ``traditional'' baseline is to use camera intrinsics and extrinsics to \emph{project} 2D detections to 3D. It can roughly localize a detected 2D box in 3D but cannot determine the object's pose and shape. We address this with two methods: (1) applying class-specific priors estimated from training data (dubbed ``proj. w/ class prior''), and (2) training a model to predict the pose\&shape for the 2D box of an object (dubbed ``proj. w/ learning''). Table~\ref{tab:benchmarking-results(0.5,0.25,0.25)} provides their results, showing that both methods underperform our Pro3D.
This demonstrates the significant role of Pro3D's 3D detection backbone, which learns to better lift 2D detections and estimate objects' pose and shape.

One may ask whether Pro3D's performance gains stem from using a powerful 2D detector or its lifting method.
First, using powerful 2D detectors does help.
We can compare FCOS3D which adds a 3D head on top of a 2D detector for 3D detection.
As the original FCOS3D exploits a Res101 backbone, we improve it by using the stronger detector DINO (with a Transformer backbone) and finetune it on the DAIR-V2X-I dataset.
We name this improved version FCOS3D+.  
Table~\ref{tab:benchmarking-results(0.5,0.25,0.25)} shows that FCOS3D+ indeed improves over FCOS3D. 
Moreover,
Fig.~\ref{fig:YOLO-v7 and more 2D annotations} justifies this further: using stronger 2D detectors improves 3D detection further.
Second, the lifting method matters, as shown from the comparison between FCOS3D+ and our Pro3D, both of which use DINO as the 2D detector.
Pro3D significantly outperforms FCOS3D+, owing to Pro3D's 3D backbone dedicated to lifting 2D detections.
Table~\ref{tab:benchmarking-results(0.5,0.25,0.25)} also shows that stronger lifting methods (BEVSpread$>$BEVHeight$>$BEVDepth) yield more gains.

\subsection{Ablation Study and Analysis}

{\bf Exploiting a scene prior improves 3D detection.} 
We concatenate a testing image and the scene prior (a ``blank'' RGB background image as shonw in Fig.~\ref{fig:background}) as a 6-channel input to the 3D detector.
Table~\ref{tab:background} shows that adding the scene prior remarkably improves roadside 3D detection.
To study how different scene prior affects the final performance, we generate scene priors over training frames captured at night.
Table~\ref{tab:results-of-priors} shows that our method is quite robust to scene priors that have different illuminations.

{\bf Exploiting a scene prior particularly improves small object detection.} 
Recall that the scene prior is a background reference.
It helps precise 2D localization of small objects by grounding them at their contacts on the ground.
This helps 3D detectors lift the precise 2D boxes  for better 3D detections.
Fig.~\ref{fig:visualcompare_scene_prior} demonstrates this by comparing  Pro3D's predictions without versus with such a scene prior.

\begin{figure}[t]
\centering
\centering
\includegraphics[width=1.0\linewidth]{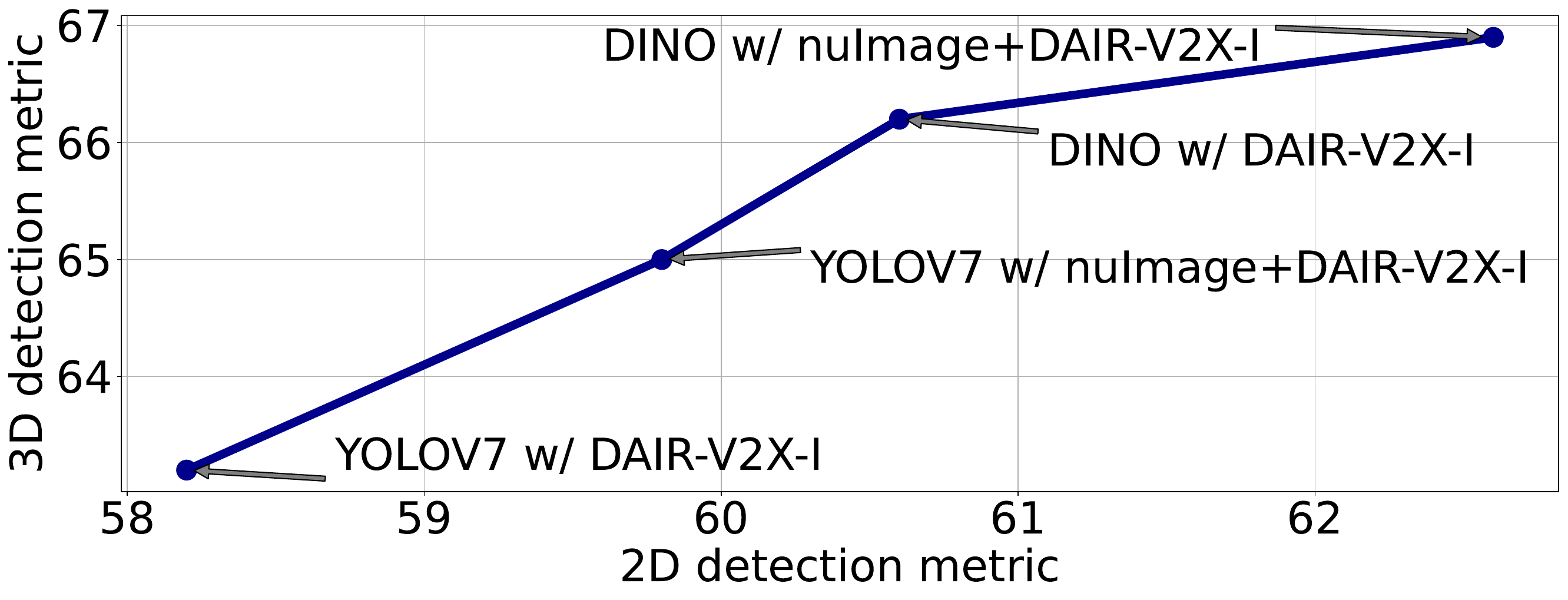}
\hfill
\vspace{-7mm}
\caption{\small
We train different 2D detectors (YOLOV7 \cite{wang2022yolov7} and DINO \cite{zhang2022dino}) on different amounts of 2D annotations (provided by  nuImages \cite{caesar2020nuscenes}) for Pro3D.
We measure the 3D detection by averaging AP over the three classes vehicle, cyclist and pedestrian.
Training on more 2D annotations leads to better 2D detectors and importantly, increases 3D detection performance of Pro3D.
}
\label{fig:YOLO-v7 and more 2D annotations}
\vspace{-1mm}
\end{figure}

{\bf Exploiting a stronger 2D detection yields better 3D detection performance.}
We train stronger 2D detectors (e.g., YOLOV7~\cite{wang2022yolov7} and DINO \cite{zhang2022dino}) on more 2D annotations from nuImages~\cite{caesar2020nuscenes}.
Fig.~\ref{fig:YOLO-v7 and more 2D annotations} plots 3D detection performance by our Pro3D versus 2D detection performance (by training different 2D detectors).
Results show that exploiting stronger 2D detectors greatly improves 3D detection performance with our Pro3D.

{
\setlength{\tabcolsep}{0.8mm}
\begin{table}[t]
\small
\centering
\caption{\small
{\bf Comparison of different prompting methods.}
As a baseline, we train BEVHeight with scene prior (Table~\ref{tab:background}).
First, all prompting methods remarkably improves over BEVHeight.
Second,
prompting with predicted 2D box coordinates (``fuse 2D centers'')  outperforms features.
Third, additionally adding the predicted labels in prompts improves further.
}
\vspace{-3mm}
\scalebox{0.7}{
\begin{tabular}{l ccc ccc ccc}
\toprule
\multirow{2}{*}{Method}            & 
\multicolumn{3}{c}{Vehicle} &\multicolumn{3}{c}{Cyclist}
&\multicolumn{3}{c}{Pedestrian} \\
\cmidrule(r){2-4} \cmidrule(r){5-7} \cmidrule(r){8-10}
& {\tt Easy}   & {\tt Mid}   & {\tt Hard}   & {\tt Easy}   & {\tt Mid}    & {\tt Hard}
      & {\tt Easy}   & {\tt Mid}    & {\tt Hard}   \\
\midrule
\rowcolor{lightgrey} \multirow{1}{*}{BEVHeight w/ scene prior}  &  \multirow{1}{*}{79.2}   &   \multirow{1}{*}{67.7}       &   \multirow{1}{*}{67.8}    &   \multirow{1}{*}{62.5}    &  \multirow{1}{*}{62.3}      &   \multirow{1}{*}{62.7}       &   \multirow{1}{*}{44.3}    &   \multirow{1}{*}{42.5}   & \multirow{1}{*}{42.7}   \\
+ concatenate feaMaps &  \multirow{1}{*}{79.6}   &   \multirow{1}{*}{68.0}       &   \multirow{1}{*}{67.7}    &   \multirow{1}{*}{63.0}    &  \multirow{1}{*}{62.6}      &   \multirow{1}{*}{62.7}       &   \multirow{1}{*}{44.2}    &   \multirow{1}{*}{42.6}   & \multirow{1}{*}{42.8}   \\

+ attentively fuse local vectors   &  
\multirow{1}{*}{79.6}   &   \multirow{1}{*}{68.1}       &   \multirow{1}{*}{68.1}    &   \multirow{1}{*}{64.0}    &  \multirow{1}{*}{63.8}      &   \multirow{1}{*}{63.9}       &   \multirow{1}{*}{44.1}    &   \multirow{1}{*}{43.3}   & \multirow{1}{*}{43.4}   \\

+ attentively fuse box centers & \multirow{1}{*}{79.8}   &   \multirow{1}{*}{68.3}       &   \multirow{1}{*}{68.3}    &   \multirow{1}{*}{65.2}    &  \multirow{1}{*}{65.0}      &   \multirow{1}{*}{65.1}       &   \multirow{1}{*}{46.8}    &   \multirow{1}{*}{45.7}   & \multirow{1}{*}{45.8}   \\

 \quad + predicted labels  &  
\multirow{1}{*}{80.2} &   
\multirow{1}{*}{68.6}       &   \multirow{1}{*}{68.6}     &   \multirow{1}{*}{66.1}    &   \multirow{1}{*}{65.5}   & \multirow{1}{*}{65.6}      &   \multirow{1}{*}{47.5}    &  \multirow{1}{*}{46.4}      &   \multirow{1}{*}{46.5}   \\

+ attentively fuse 2D boxes & \multirow{1}{*}{80.1}   &   \multirow{1}{*}{68.6}       &   \multirow{1}{*}{68.5}    &   \multirow{1}{*}{66.7}    &  \multirow{1}{*}{66.1}      &   \multirow{1}{*}{66.2}       &   \multirow{1}{*}{48.2}    &   \multirow{1}{*}{47.1}   & \multirow{1}{*}{47.2}   \\

\quad + predicted labels  &  
\multirow{1}{*}{\textbf{80.5}} &   
\multirow{1}{*}{\textbf{68.9}}       &   \multirow{1}{*}{\textbf{68.9}}     &   \multirow{1}{*}{\textbf{68.9}}    &   \multirow{1}{*}{\textbf{68.0}}   & \multirow{1}{*}{\textbf{68.2}}      &   \multirow{1}{*}{\textbf{50.7}}    &  \multirow{1}{*}{\textbf{50.0}}      &   \multirow{1}{*}{\textbf{50.1}}   \\
\quad  \emph{performance gain} &  \multirow{1}{*}{\color{darkgreen}+1.3}   &   \multirow{1}{*}{\color{darkgreen}+1.2}       &   \multirow{1}{*}{\color{darkgreen}+1.1}    &   \multirow{1}{*}{\color{darkgreen}+6.4}    &  \multirow{1}{*}{\color{darkgreen}+5.7}      &   \multirow{1}{*}{\color{darkgreen}+5.5}       &   \multirow{1}{*}{\color{darkgreen}+6.4}    &   \multirow{1}{*}{\color{darkgreen}+7.5}   & \multirow{1}{*}{\color{darkgreen}+7.4}   \\
\bottomrule
\end{tabular}
}
\vspace{-3mm}
\label{tab:results-of-different-prompt-information}
\end{table}
}

{\bf What to use as prompts?} 
We study what information from the 2D detector can be used as prompts to facilitate 3D detector training,
including the intermediate feature map, the feature vectors located at a 2D detection, and even the simplistic 2D detection output (i.e., predicted box coordinates and class label).
They require different fusion methods to accommodate their different dimensions.
For example, when using feature maps, one can concatenate them with the feature map of 3D detector backbone,
while using feature vectors or detections require fusing vectors with the feature maps.
Hence, we use concatenation to fuse feature maps, and an attention module to fuse the feature vectors.
Table~\ref{tab:results-of-different-prompt-information} compares them.
Surprisingly, using the simple detections (2D boxes and class labels) as prompts performs the best among the three methods: 
it achieves [68.5, 66.2, 47.2] AP for {\tt Hard} examples of the three classes [vehicle, cyclist, pedestrian].
Furthermore, additionally incorporating the predicted class labels in prompts boosts the performance to [68.9, 68.2, 50.1] AP on {\tt Hard} examples of  the three classes!
In hindsight, fusing box coordinates and labels provides both semantic and spatial information for the 3D detector to leverage, facilitating 3D detection. In contrast, fusing feature maps does not pinpoint the location of the objects and hence does not help  3D detector training.

{
\setlength{\tabcolsep}{1.0mm}
\begin{table}[t]
\small
\centering
\caption{\small
{\bf Stage-wise training (SW) vs. joint training (JT)  for Pro3D.}
We report AP metrics at IoU=0.5, 0.25, 0.25 in BEV on the DAIR-V2X-I benchmark.
As a baseline, we train Pro3D with BEVHeight and scene prior.
Stage-wise training allows us to train separate 2D and 3D detector backbones, i.e., their backbones are not shared (NS).
This allows us to exploit more 2D annotated data (MD) to train better 2D detectors, e.g., using nuImages~\cite{caesar2020nuscenes} to pretrain a 2D detector.
Results convincingly show that SW outperforms JT for roadside monocular 3D detection.
With NS and MD, Pro3D yields more notable improvements.
}
\vspace{-3mm}
\scalebox{0.78}{
\begin{tabular}{l ccc ccc ccc}
\toprule
\multirow{2}{*}{Method}            & 
\multicolumn{3}{c}{Vehicle} &\multicolumn{3}{c}{Cyclist}
&\multicolumn{3}{c}{Pedestrian} \\
\cmidrule(r){2-4} \cmidrule(r){5-7} \cmidrule(r){8-10}
& {\tt Easy}   & {\tt Mid}   & {\tt Hard}   & {\tt Easy}   & {\tt Mid}    & {\tt Hard}
      & {\tt Easy}   & {\tt Mid}    & {\tt Hard}   \\
\midrule
\rowcolor{lightgrey} Pro3D  &   80.5       &   68.9    &   68.9    &  68.9      &   68.0      &   68.2    &   50.7  &  50.0 &  50.1   \\


\quad + JT   &   80.6       &   68.8    &   68.9    &  68.9      &   68.2       &   68.3    &   50.6  &  50.1  &  50.3   \\
\quad + SW   &    81.5      &   69.3    &   69.6    &  69.5     &   69.1      &   69.2     &   51.3  &  50.7 &  50.3   \\
\quad + SW, NS   &    82.3     &   69.7    &   70.0    &  70.0      &   69.9      &   70.1    &   52.1  &  51.0  &  50.8   \\
\quad + SW, NS, MD   &    \textbf{82.7}      &   \textbf{70.1}    &   \textbf{70.3}    &  \textbf{70.5}     &   \textbf{70.4}      &   \textbf{70.6}    &   \textbf{52.8}  &  \textbf{51.7}  &  \textbf{51.3}   \\
\quad \emph{perf. gain}  &  \multirow{1}{*}{\color{darkgreen}+2.2}   &   \multirow{1}{*}{\color{darkgreen}+1.2}       &   \multirow{1}{*}{\color{darkgreen}+1.4}    &   \multirow{1}{*}{\color{darkgreen}+1.6}    &  \multirow{1}{*}{\color{darkgreen}+2.4}      &   \multirow{1}{*}{\color{darkgreen}+2.4}       &   \multirow{1}{*}{\color{darkgreen}+2.1}    &   \multirow{1}{*}{\color{darkgreen}+1.7}   & \multirow{1}{*}{\color{darkgreen}+1.2}   \\
\bottomrule
\end{tabular}
}
\vspace{-4mm}
\label{tab:ablation-joint-training}
\end{table}
}

{\bf Stage-wise training vs. joint training.} 
We test different training paradigms in our Pro3D, e.g.,
(1) joint end-to-end training the 3D detector using both 2D and 3D losses,
and (2) stage-wise training the 2D detector first and then the 3D detector.
Note that stage-wise training allows us to train separate 2D and 3D detector backbones, i.e., their backbones are \emph{n}ot \emph{s}hared (NS).
This further makes it convenient to exploit \emph{m}ore 2D annotated \emph{d}ata (MD) to train better 2D detectors, e.g., using nuImages~\cite{caesar2020nuscenes} to pretrain a 2D detector.
Results in Table~\ref{tab:ablation-joint-training} show that stage-wise training outperforms joint training for roadside monocular 3D detection.

{\bf How to train 3D detectors with 2D detection prompts?}
To train the 3D detector, we can either use predicted 2D boxes or the
ground-truth as  prompts. 
Table~\ref{tab:results of different training methods of Pro3D} compares their performance w.r.t AP in BEV on the DAIR-V2X-I benchmark,
showing that the adpoting the former yields better performance.

{
\setlength{\tabcolsep}{1.5mm}
\begin{table}[t]
\small
\centering
\caption{\small
{\bf How to train Pro3D}.
When training the 3D detector,
we can construct prompts as either ground-truth 2D boxes (grnd) or predicted ones (pred).
We compare them on DAIR-V2X-I w.r.t AP in BEV.
Results show that using latter performs better.
}
\vspace{-3mm}
\scalebox{0.8}{
\begin{tabular}{l ccc ccc ccc}
\toprule
\multirow{2}{*}{Method}            & 
\multicolumn{3}{c}{Vehicle} &\multicolumn{3}{c}{Cyclist}
&\multicolumn{3}{c}{Pedestrian} \\
\cmidrule(r){2-4} \cmidrule(r){5-7} \cmidrule(r){8-10} 
& {\tt Easy}   & {\tt Mid}   & {\tt Hard}   & {\tt Easy}   & {\tt Mid}    & {\tt Hard}
      & {\tt Easy}   & {\tt Mid}    & {\tt Hard}   \\
\midrule
w/ grnd  &    82.7      &   70.1    &   70.3    &  70.5     &   70.4      &   70.6    &   52.8  &  51.7  &  51.3   \\


w/ pred  &  \multirow{1}{*}{\textbf{82.9}}   &   \multirow{1}{*}{\textbf{70.7}}       &   \multirow{1}{*}{\textbf{70.8}}  &   \multirow{1}{*}{\textbf{71.3}}    &   \multirow{1}{*}{\textbf{71.2}}   & \multirow{1}{*}{\textbf{71.4}}    &   \multirow{1}{*}{\textbf{53.9}}    &  \multirow{1}{*}{\textbf{52.9}}      &   \multirow{1}{*}{\textbf{52.6}}   \\

\bottomrule
\end{tabular}
}
\vspace{-1mm}
\label{tab:results of different training methods of Pro3D}
\end{table}
}

{
\setlength{\tabcolsep}{0.6em}
\begin{table}[t] 
\centering
 \caption{\small
{\bf  Comparison of inference time and parameters between  Pro3D and BEVHeight.} 
AP means the average AP in BEV over all the three classes in DAIR-V2X-I; ``\#param'' counts the parameters (in million); time is the wall-clock in ms, FPS is frames per second.
We use two different 2D detectors in our Pro3D, namely YOLOV7 and DINO.
We list the number of parameters of 3D detector plus 2D detector.
Our Pro3D ``w/ YOLOV7'' boosts 3D detection performance by 7.9 AP over the prior art BEVHeight, with faster inference speed (15.4 vs. 13.0 FPS) and fewer  parameters (86.6M vs. 94.7M)!
}
 \vspace{-3mm}
\scalebox{0.8}{
\begin{tabular}{lcccc}
\toprule 
Methods
& AP
& \#param.
& time 
& FPS \\
\midrule
BEVHeight \cite{yang2023BEVHeight} 
& 57.1
& {94.7\ \ \  \ \ \ \ \ \ }
& 77 
& 13.0 \\
{\bf Pro3D} w/ BEVHeight + YOLOV7 
& 65.0
& \textbf{82.1+4.5 \ }
& \textbf{65}  
& \textbf{15.4} \\
{\bf Pro3D} w/ BEVHeight + DINO 
& \textbf{66.9}
& {82.1+21.2}
& 73 
& 13.7 \\
\bottomrule
\end{tabular}
}
\vspace{-5mm}
\label{tab:inference-time} 
\end{table}
}

{\bf Latency, parameters, and training time.} 
Recall that Pro3D achieves the best performance by using separate 2D and 3D detectors without sharing the visual encoder (Table~\ref{tab:ablation-joint-training}).
We study this model in Table~\ref{tab:inference-time} w.r.t model parameters, the wall-clock time, and FPS in processing a single RGB frame, as well as training time.
For inference, we run the 2D detector (either YOLOV7 or DINO) and the 3D detector encoder in parallel.
It is worth noting that Pro3D does not require an FPN for proposal detection as the 2D detector provides pinpointed detections already; it also replaces Res18  encoder (as adopted in prior methods~\cite{yang2023BEVHeight}) with a single convolution layer (Sec.~\ref{ssec:setting}).
These changes make our Pro3D faster than the original BEVHeight in inference, regardless exploiting either YOLOV7 \cite{wang2022yolov7}  or DINO \cite{yang2023BEVHeight} as the 2D detector.
The changes also accelerate training.
Table~\ref{tab:training_time} compares the training times (in minutes) of BEVHeight and our Pro3D (built on BEVHeight). 
One might have thought that Pro3D requires more training time as it finetunes a 2D detector (YOLOv7 or DINO); in contrast, it needs less training time: a finetuned 2D detector provides a better base model for 3D training and thus reduces the total epochs needed.
Further, owing to the modification on BEVHeight for Pro3D (removing FPN, replacing MLP head with a single convolution layer), Pro3D requires less training time in each epoch.
In summary, Pro3D not only achieves better performance than BEVHeight, but also has faster inference and training speed.

{
\setlength{\tabcolsep}{1.2mm}
\begin{table}[t]
\small
\centering
\caption{\small
{\bf  Comparison of training time between  Pro3D and BEVHeight.} 
Following Table~\ref{tab:inference-time}, we build Pro3D with BEVHeight for fair comparison.
One might think, as Pro3D finetunes 2D detectors (YOLOv7 or DINO), it should need more training time. Quite differently, a finetuned 2D detector provides a better base model for 3D training and thus reduces the total epochs needed. As Pro3D does not need an FPN and replaces the BEVHeight's Res18-based BEV encoder with a single convolution layer, it needs less time in each epoch.
}
\vspace{-3mm}
\scalebox{0.82}{
\begin{tabular}{lcccr}
\toprule
Method & {2D det. ft.} & {min/epoch} & {\#epochs} & {total time in minutes} \\
\midrule
BEVHeight & n/a & 13 & 75 & $13 \times 75 = 975$ \\
Pro3D + YOLOV7  & 35 & 11 & 40 & $35 + 11 \times 40 = 475$ \\
Pro3D + DINO  & 60 & 11 & 40 & $60 + 11 \times 40 = 500$ \\
\bottomrule
\end{tabular}
}
\vspace{-4mm}
\label{tab:training_time}
\end{table}
}

\subsection{Societal Impacts and Limitations}

{\bf Societal impacts.}
We note potential risks if directly using the proposed Pro3D in real-world applications such as vehicle-vehicle communication and vehicle-infrastructure cooperative perception. 
These applications should evaluate our approach in their whole system w.r.t safety and robustness.
Moreover, as methods (including our Pro3D) developed with existing datasets might be biased towards specific cities and countries.
deploying Pro3D requires caution.

{\bf Limitations and future work.}
Currently, our Pro3D exploits a scene prior learned before hand;
the change of camera pose (e.g., due to camera replacement or strong wind) can affect performance.
While we have studied the robustness of Pro3D to scene priors generated in daytime or at night (Table~\ref{fig:visualcompare_scene_prior}), we did not extend this study to different weather conditions as current datasets did not capture images under diverse weathers.
Moreover, Pro3D outputs a 3D detection given a 2D detection.
As 2D detections can be false positives, doing so inevitably hallucinates 3D detections which certainly are incorrect.
Future work will analyze these aspects in more depth.

\section{Conclusion}

We address the problem of roadside monocular 3D detection and propose a novel design, Promptable 3D detector (Pro3D), which leverages a pre-trained 2D detector to facilitate 3D detector training.
Our Pro3D runs a 2D detector to obtain 2D detections,
and uses them to prompt the 3D detector to make 3D predictions.
We present novel modules to encode 2D detections and fuse them with features from the 3D detector's backbone.
Moreover,
we present a simple method to obtain a scene prior, by incorporating which our Pro3D achieves  further improvements.
We validate Pro3D through extensive experiments and ablation studies.
Comprehensive results show that Pro3D significantly outperforms prior works on publicly available benchmarks.




\bibliography{references}
\bibliographystyle{ieeenat_fullname}

\appendix


\clearpage
\setcounter{page}{1}
\maketitlesupplementary

\renewcommand{\thesection}{\Alph{section}}
\setcounter{section}{0} 

\section*{}
\begin{center}
    \emph{\bf \em \large Outline}
\end{center}
This document supplements the main paper with further details of Pro3D and more visualizations. Below is the outline of this document.
\begin{itemize}[topsep=-3pt, partopsep=3pt, leftmargin=0.3in]
\item 
    {\bf Section \ref{sec:detail-implementation-for-modules}.} We provide implementation details of the prompt encoder in Pro3D.
\item 
    {\bf Section \ref{sec:details-class-grouping-appendix}.} We discuss our loss design in Pro3D.
\item 
    {\bf Section \ref{sec:occlusion-truncation}.} 
    We evaluate models' performance under occlusion and truncation scenarios.
\item 
    {\bf Section \ref{sec:robustness}.} We study the robustness of Pro3D to camera noises.

\item  {\bf Section \ref{sec:2D-metrics-by-3D-detectors}.} We report 2D metrics by roadside monocular 3D detectors.

\item  {\bf Section \ref{sec:cross-dataset-evaluation}.} We conduct cross-dataset evaluation.

\item 
    {\bf Section \ref{sec:inference time}.} 
    We compare the inference speed of Pro3D using different 3D detectors.

\item 
    {\bf Section \ref{sec:Demo-code}.} 
    We provide our code.
    
\item
    {\bf Section \ref{sec:more-visualizations}.} 
    We provide more visual results, including demo videos, and a Jupyter Notebook of code.

\end{itemize}


\section{Details of Prompt Encoder}
\label{sec:detail-implementation-for-modules}
The prompt encoder transforms the 2D detection outputs to features through three steps: (1) normalize the top-left and bottom-right coordinates of a 2D detection box, denoted as $A \in \RB^{2\times2}$;
(2) multiply $A$ by a random matrix $B \in \RB^{2\times512}$ whose entries are initialized using a Gaussian distribution, and then add a learnable matrix $C \in \RB^{2\times512}$ to get $D = A\times B + C$;
(3) repeat the predicted class label (represented as a unique class ID) 512 times to get a 512-dimensional vector, and concatenate it with $D$ to obtain the final feature prompts $E \in \RB^{3\times 512}$.

\begin{figure}[t]
\centering
\centering
\includegraphics[trim=0cm 0 0cm 0cm, clip, width=1.02\linewidth]{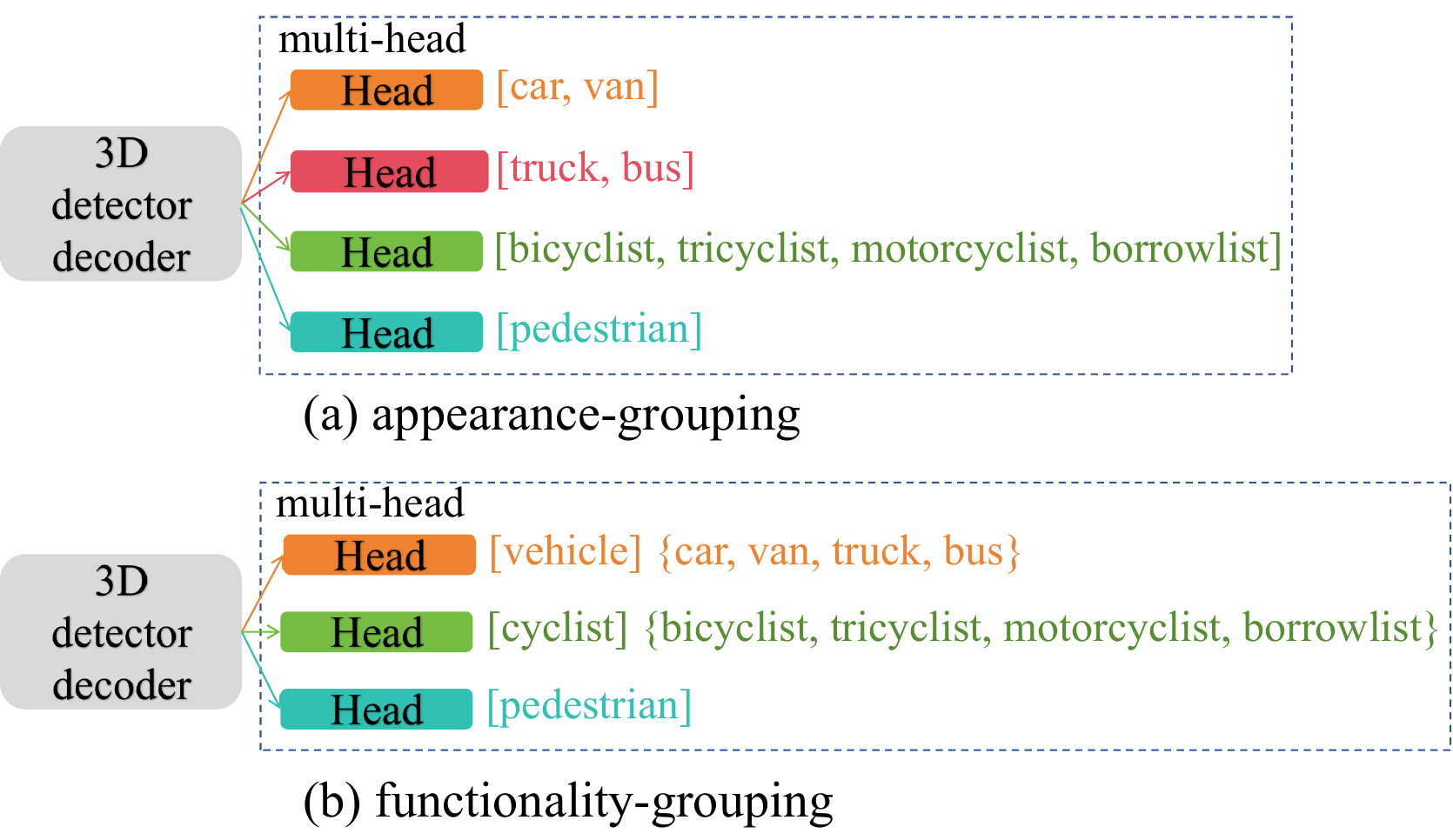}
\vspace{-7mm}
\caption{\small
Comparison of two grouping methods in the loss design:
(a) appearance-based grouping, and (b) functionality-based grouping. 
The former is commonly adopted in existing roadside 3D detectors that group classes according to their appearance similarity and use multi-head for prediction (each head is responsible for $K$-way fine-grained prediction.
The latter is our proposed method that groups classes according to their functionalities and uses multi-head for prediction -- each head is a single super-class predictor, which merges \{{\em car}, {\em van}, {\em truck}, {\em bus}\}  into {\em vehicle}, and  \{{\em bicyclist}, {\em tricyclist}, {\em motorcyclist}, {\em barrowlist}\} into {\em cyclist}.
}
\label{fig:3D detector with different grouping methods}
\end{figure}

{\setlength{\tabcolsep}{3.8mm}
\begin{table*}
    \caption{\small
    \textbf{Results of occluded and truncated obstacles}. We split occlusions and truncations with different ranges (0\%$\sim$50\% and 50\%$\sim$100\%), respectively.
    Our approach Pro3D resoundingly outperforms BEVHeight \cite{yang2023BEVHeight} and BEVSpread \cite{wang2024bevspread} on all the three superclasses in each range.}
    \label{tab:occlusion-truncation}
    \vspace{-2mm}
    \begin{subtable}[t]{0.48\textwidth}
        \centering
        \caption{Occlusion of 0\%$\sim$50\%}
        \vspace{-1mm}
        \label{tab:occlusion1}
            \scalebox{0.85}{
            \begin{tabular}{lcccc}
            \toprule
            Methods
            & Vehicle 
            & Cyclist 
            & Pedestrian  \\
            \midrule
            BEVHeight 
            & 75.6
            & 66.8
            & 48.9  \\
    {\bf Pro3D} w/ BEVHeight  
            & 80.8
            & 76.6
            & 58.0  \\
    \midrule
            BEVSpread 
            & 77.5
            & 67.7
            & 51.2  \\
    {\bf Pro3D} w/ BEVSpread  
            & \textbf{83.2}
            & \textbf{76.8}
            & \textbf{60.1} \\
            \bottomrule
            \vspace{1mm}
            \end{tabular}
            }
    \end{subtable}
    \begin{subtable}[t]{0.48\textwidth}
        \centering
        \caption{Occlusion of 50\%$\sim$100\%}
        \vspace{-1mm}
        \label{tab:occlusion2}
            \scalebox{0.85}{
            \begin{tabular}{lcccc}
            \toprule
            Methods
            & Vehicle 
            & Cyclist 
            & Pedestrian  \\
            \midrule
            BEVHeight 
            & 51.2
            & 36.9
            & 19.2  \\
        {\bf Pro3D} w/ BEVHeight  
            & 58.0
            & 42.3
            & 28.1 \\
            \midrule        
            BEVSpread
            & 53.3
            & 38.6
            & 19.9  \\
        {\bf Pro3D} w/ BEVSpread 
            & \textbf{60.1}
            & \textbf{44.6}
            & \textbf{28.5}  \\
            \bottomrule
            \end{tabular}
            }
    \end{subtable}
    \begin{subtable}[t]{0.48\textwidth}
        \centering
        \caption{Truncation of 0\%$\sim$50\%}
        \vspace{-1mm}
        \label{tab:truncation1}
            \scalebox{0.85}{
            \begin{tabular}{lcccc}
            \toprule
            Methods
            & Vehicle 
            & Cyclist 
            & Pedestrian  \\
            \midrule
            BEVHeight 
            & 76.2
            & 68.0
            & 50.3  \\
            {\bf Pro3D} w/ BEVHeight  
            & 83.9
            & 76.8
            & 60.0  \\  
            \midrule
            BEVSpread 
            & 78.8
            & 69.3
            & 52.6  \\
            {\bf Pro3D} w/ BEVSpread
            & \textbf{85.0}
            & \textbf{77.9}
            & \textbf{61.6}  \\
            \bottomrule
            \end{tabular}
            }
    \end{subtable}
    \hfill
    \begin{subtable}[t]{0.48\textwidth}
        \centering
        \caption{Truncation of 50\%$\sim$100\%}
        \vspace{-1mm}
        \label{tab:truncation2}
            \scalebox{0.85}{
            \begin{tabular}{lcccc}
            \toprule
            Methods
            & Vehicle 
            & Cyclist 
            & Pedestrian  \\
            \midrule
            BEVHeight 
            & 38.7
            & 18.6
            & 10.8  \\
            {\bf Pro3D} w/ BEVHeight  
            & 45.5
            & 27.3
            & 20.5 \\  
            \midrule
            BEVSpread
            & 38.9
            & 19.1
            & 11.2  \\
            {\bf Pro3D} w/ BEVSpread
            & \textbf{45.6}
            & \textbf{27.6}
            & \textbf{20.6}  \\
            \bottomrule
            \end{tabular}
            }
    \end{subtable}
\end{table*}
}

{
\setlength{\tabcolsep}{1.5mm}
\begin{table}
\small
\centering
\caption{\small
{\bf Ablation on class-grouping strategies} (w.r.t metric AP in BEV).
Grouping classes differently with multi-head design (each detector head makes predictions within a specific superclass) makes a difference in training and yields detectors performing differently. 
Briefly, instead of appearance-based grouping (AG) which groups classes w.r.t object appearance as done by BEVHeight, we do functionality-based grouping (FG), e.g., merging \{{\em car}, {\em van}, {\em truck}, {\em bus}\} into the superclass {\em vehicle}.
To justify the superiority of FG over AG, we replace the original AG of BEVHeight \cite{yang2023BEVHeight} and BEVSpread \cite{wang2024bevspread} with FG, particularly improving on the vehicle class (cf. ``BEVHeight w/ FG'' vs. ``BEVHeight w/ AG'', ``BEVSpread w/ FG'' vs. ``BEVSpread w/ AG'').
}
\vspace{-2mm}
\scalebox{0.72}{
\begin{tabular}{l ccc ccc ccc}
\toprule
\multirow{2}{*}{Method}            & 
\multicolumn{3}{c}{Vehicle} &\multicolumn{3}{c}{Cyclist}
&\multicolumn{3}{c}{Pedestrian} \\
\cmidrule(r){2-4} \cmidrule(r){5-7} \cmidrule(r){8-10}
& {\tt Easy}   & {\tt Mid}   & {\tt Hard}   & {\tt Easy}   & {\tt Mid}    & {\tt Hard}
      & {\tt Easy}   & {\tt Mid}    & {\tt Hard}   \\
\midrule
BEVHeight w/ AG  &  \multirow{1}{*}{77.8}   &   \multirow{1}{*}{65.8}       &   \multirow{1}{*}{65.9}   &   \multirow{1}{*}{60.2}    &   \multirow{1}{*}{60.1}   & \multirow{1}{*}{60.5}  &   \multirow{1}{*}{41.2}    &  \multirow{1}{*}{39.3}      &   \multirow{1}{*}{39.5}   \\
BEVHeight w/ FG
& \multirow{1}{*}{80.0}   &   \multirow{1}{*}{67.9}       &   \multirow{1}{*}{68.1}    &   \multirow{1}{*}{60.5}    &  \multirow{1}{*}{60.3}      &   \multirow{1}{*}{60.7}       &   \multirow{1}{*}{41.4}    &   \multirow{1}{*}{39.4}   & \multirow{1}{*}{39.7}   \\
\textit{performance gain} &  \multirow{1}{*}{\color{darkgreen}+2.2}   &   \multirow{1}{*}{\color{darkgreen}+2.1}       &   \multirow{1}{*}{\color{darkgreen}+2.2}    &   \multirow{1}{*}{\color{darkgreen}+0.3}    &  \multirow{1}{*}{\color{darkgreen}+0.2}      &   \multirow{1}{*}{\color{darkgreen}+0.2}       &   \multirow{1}{*}{\color{darkgreen}+0.2}    &   \multirow{1}{*}{\color{darkgreen}+0.1}   & \multirow{1}{*}{\color{darkgreen}+0.2}   \\
\midrule
BEVSpread w/ AG &  \multirow{1}{*}{79.1}   &   \multirow{1}{*}{66.8}       &   \multirow{1}{*}{66.9}    &   \multirow{1}{*}{62.6}    &   \multirow{1}{*}{63.5}   & \multirow{1}{*}{63.8}   &   \multirow{1}{*}{46.5}    &  \multirow{1}{*}{44.5}      &   \multirow{1}{*}{44.7}       \\
BEVSpread w/ FG    
&  \multirow{1}{*}{81.1}   &   \multirow{1}{*}{68.8}       &   \multirow{1}{*}{69.0}  &   \multirow{1}{*}{62.7}    &   \multirow{1}{*}{63.7}   & \multirow{1}{*}{63.9}   &   \multirow{1}{*}{46.6}    &  \multirow{1}{*}{44.7}      &   \multirow{1}{*}{44.8} \\
\textit{performance gain} &  \multirow{1}{*}{\color{darkgreen}+2.0}   &   \multirow{1}{*}{\color{darkgreen}+2.0}       &   \multirow{1}{*}{\color{darkgreen}+2.1}    &   \multirow{1}{*}{\color{darkgreen}+0.1}    &  \multirow{1}{*}{\color{darkgreen}+0.2}      &   \multirow{1}{*}{\color{darkgreen}+0.1}       &   \multirow{1}{*}{\color{darkgreen}+0.1}    &   \multirow{1}{*}{\color{darkgreen}+0.2}   & \multirow{1}{*}{\color{darkgreen}+0.1}   \\
\bottomrule
\end{tabular}
}
\vspace{-1mm}
\label{tab:results of different prompt classes and detection heads of Pro3D}
\end{table}
}

\section{Loss Design}
\label{sec:details-class-grouping-appendix}

Class grouping is a common strategy in 3D object detector training.
It merges classes into some superclasses, allowing trained features to be shared within superclasses \cite{zhu2019class}.
Moreover, multi-head design follows along with groups of classes for discriminative predictions across fine-grained classes \cite{yin2021center}.
BEVHeight \cite{yang2023BEVHeight} and BEVSpread \cite{wang2024bevspread} adopt class grouping and multi-head design by considering the similarities among object appearance, e.g., size and shape in the 3D world.
For example, on DAIR-V2X-I, they create three superclasses that merge \{\emph{truck}, \emph{bus}\}, \{\emph{car}, \emph{van}\}, and \{\emph{bicyclist}, \emph{tricyclist}, \emph{motorcyclist},  \emph{barrowlist}\}, respectively, along with the origin \emph{pedestrian} class.
The rationale of such grouping is based on object appearance, e.g., both {\em truck} and {\em bus} have large size. We call this grouping strategy \emph{appearance-based grouping}.
With such a group strategy, they learn a four-head detector with each making predictions for the corresponding superclass.
Differently,
we adopt \emph{functionality-based grouping}, which merges
\{{\em car}, {\em van}, {\em truck}, {\em bus}\} 
into {\em vehicle}, and 
\{{\em bicyclist}, {\em tricyclist}, {\em motorcyclist}, {\em barrowlist}\} into {\em cyclist}.
This is motivated by the fact that vehicles, cyclists and pedestrians appear at relatively fixed regions on the roadside images.
We expect this grouping strategy, or using the location prior of different classes, to facilitate training 3D object detectors on roadside images.
Fig.~\ref{fig:3D detector with different grouping methods} compares the two grouping methods.
Perhaps from the same functionality perspective, the DAIR-V2X-I benchmark also merges classes into such superclasses for evaluation.
For fair comparison, we use our functionality-based grouping loss to train the 3D detector (e.g., BEVHeight or BEVSpread)  within our Pro3D framework to compare with the original 3D detectors.
Our extensive study shows that trainig with the functionality-based grouping loss remarkably improves detection performance (Table \ref{tab:results of different prompt classes and detection heads of Pro3D}).

\section{Analysis of Occlusion and Truncation}
\label{sec:occlusion-truncation}
Objects might be occluded by others or truncated around the image borders.
We carry out breakdown analysis how detectors perform in these scenarios.
The DAIR-V2X-I dataset provides occlusion and truncation tags. 
We split occlusions and truncations with 0\%$\sim$50\% and 50\%$\sim$100\% for different superclasses on \cite{yu2022dair}, respectively. As shown in Table~\ref{tab:occlusion-truncation}, our method gets better APs on the all superclasses in different ranges of  occlusions and truncations.
This shows that our method Pro3D is more robust than BEVHeight \cite{yang2023BEVHeight} and BEVSpread \cite{wang2024bevspread}.

\section{Robustness to Noises of Camera Pose}
\label{sec:robustness}
We consider the robustness of Pro3D to noises in camera calibration parameters.
We follow BEVHeight \cite{yang2023BEVHeight} to add noises on  camera pitch and roll. Table~\ref{tab:pitch_roll} compares the robustness of BEVHeight \cite{yang2023BEVHeight}, BEVSpread \cite{wang2024bevspread} and our Pro3D.
Clearly, Pro3D is more robust than prior works to camera noises.

{
\setlength{\tabcolsep}{1.6mm}
\begin{table}[t]
\small
\centering
\caption{\small
\textbf{Robustness to camera noise}.
We follow BEVHeight \cite{yang2023BEVHeight} to simulate calibration errors by adding noises to camera pitch and roll.
Results show that Pro3D is more robust than BEVHeight and BEVSPread \cite{wang2024bevspread} to these camera pose noises.}
\vspace{-2mm}
\scalebox{0.64}{
\begin{tabular}{l ccc ccc ccc}
\toprule
\multirow{2}{*}{Method}            & 
\multicolumn{3}{c}{Vehicle} &\multicolumn{3}{c}{Cyclist}
&\multicolumn{3}{c}{Pedestrian} \\
\cmidrule(r){2-4} \cmidrule(r){5-7} \cmidrule(r){8-10}

& {\tt Easy}   & {\tt Mid}   & {\tt Hard}   & {\tt Easy}   & {\tt Mid}    & {\tt Hard}
      & {\tt Easy}   & {\tt Mid}    & {\tt Hard}   \\
\midrule

BEVHeight   &   77.8       &   65.8    &   65.9       &   60.2  &  60.1  & 60.5  &  41.2      &   39.3   &   39.5\\
\quad + noise   &   62.5       &   49.9    &   50.7       &   45.4  &  43.4  & 43.8  &  34.5      &   32.6   &   32.7 \\
\quad  \textit{performance degradation} &   -15.3       &   -15.9    &   -15.2    &  -14.8      &   -16.7       &   -16.7    &   -6.7  & -6.7  & -6.8   \\
\midrule

{\bf Pro3D} w/ BEVHeight  &   83.7       &   71.8    &   72.0    &  71.3      &   71.2       &   71.4    &   53.9  &  52.9  &  52.6   \\
\quad + noise   &   72.5       &   59.9    &   60.3       &   59.0  &  60.0  & 61.9  &  47.9      &   46.8   &   46.5 \\
\quad  \textit{performance degradation} &   -11.2       &   -11.9   &   -11.7    &  -12.3      &   -11.2       &   -9.5    &  -6.0  & -6.1  & -6.1   \\

\midrule

BEVSpread   &   79.1       &   66.8    &   66.9       &   62.6  &  63.5  & 63.8  &  46.5      &   44.5   &   44.7 \\
\quad + noise   &   64.2       &   51.6    &   51.4       &   48.4  &  48.8  & 49.4  &  40.3      &   37.6   &   38.6 \\
\quad  \textit{performance degradation} &   -14.9       &   -15.2    &   -15.5    &  -14.2      &   -14.7       &   -14.4    &   -6.2  &  -6.9  & -6.1   \\


\midrule
{\bf Pro3D} w/ BEVSpread   &    85.2       &   73.1   &   73.6       &   73.1    &   73.0  &  73.1  & 55.1    &  54.3      &   54.1   \\

\quad + noise   &   73.1       &   60.3    &   61.7       &   60.4  &  61.8  & 63.1  &  48.9      &   48.0   &   47.8 \\

\quad  \textit{performance degradation}&   -12.1      &   -12.8    &   -11.9    &  -12.7     &   -11.2       &   -10.0    &   -6.2  &  -6.3  & -6.3   \\

\midrule
\end{tabular}
}
\label{tab:pitch_roll} 
\vspace{-1mm}
\end{table}
}

{
\setlength{\tabcolsep}{1.6mm}
\begin{table}[t]
\small
\centering
\caption{\small
{\bf Comparison of inference time (\emph{ms}) between different methods.}
BEVHeight~\cite{yang2023BEVHeight} has the fastest inference time comparing with other three methods.
Here, we build our Pro3D using it as the 3D detector.
It is worth noting that we adopt the functionality-based grouping loss, which induces less parameters compared to the wide-used appearance-based grouping loss (Section~\ref{sec:details-class-grouping-appendix}). This makes our Pro3D obtain the fastest inference speed among these methods.
}
\vspace{-2mm}
\scalebox{0.8}{
\begin{tabular}{l ccccc}
\toprule
\multirow{1}{*}{Method} &   
\multicolumn{1}{c}{BEVDepth}& 
\multicolumn{1}{c}{BEVHeight} &\multicolumn{1}{c}{BEVHeight++}
&\multicolumn{1}{c}{BEVSpread}&\multicolumn{1}{c}{Pro3D} \\
\midrule
time(ms) &  \multirow{1}{*}{82} & \multirow{1}{*}{77}   &   \multirow{1}{*}{92}       &   \multirow{1}{*}{90} &   \multirow{1}{*}{\textbf{65}} \\

\bottomrule
\end{tabular}
}
\vspace{-1mm}
\label{tab:inference time}
\end{table}
}

\section{2D Detection Performance of 3D Detectors}
\label{sec:2D-metrics-by-3D-detectors}

Table~\ref{tab:benchmarking-results of 2D bbox} reports 2D metrics of both the 2D detector DINO and roadside 3D detection methods including BEVDepth, BEVHeight, BEVSpread, and their improved versions by our Pro3D.
Results demonstrate that (1) existing 3D detectors (BEVDepth, BEVHeight, and BEVSpread) produce significantly lower 2D metrics than the 2D detector DINO;
(2) by exploiting DINO's 2D detections as prompt, our Pro3D greatly enhances 3D detection (see results in Table~\ref{tab:benchmarking-results(0.5,0.25,0.25)} and \ref{tab:benchmarking-results of Rope3d(0.7,0.5,0.5)} in the main paper).

{
\setlength{\tabcolsep}{0.7mm}
\begin{table}[t]
\small
\centering
\caption{\small
{\bf Comparison w.r.t 2D detection metric (mAP) on DAIR-V2X-I.}
For both 2D detector (DINO) and 3D detectors BEVDepth and BEVHeight, we train them on the training set of DARI-V2X-I.
To evaluate 3D detectors w.r.t 2D metrics, we project their 3D detections onto the 2D image plane to derive their predicted 2D boxes.
Compared to BEVDepth and BEVHeight,
DINO achieves $>$10 mAP higher on all classes and $\sim$20 mAP higher  on Pedestrian!
Owing to the effective leverage of the 2D detector DINO, our Pro3D significantly boosts performance of the compared methods.
}
\vspace{-3mm}
\scalebox{0.75}{
\begin{tabular}{l ccc ccc ccc}
\toprule
\multirow{2}{*}{Method}            & 
\multicolumn{3}{c}{Vehicle} &\multicolumn{3}{c}{Cyclist}
&\multicolumn{3}{c}{Pedestrian} \\
\cmidrule(r){2-4} \cmidrule(r){5-7} \cmidrule(r){8-10}

& {\tt Easy}   & {\tt Mid}   & {\tt Hard}   & {\tt Easy}   & {\tt Mid}    & {\tt Hard}
      & {\tt Easy}   & {\tt Mid}    & {\tt Hard}   \\
\midrule
\cellcolor{lightgrey}2D detector (DINO) \cite{zhang2022dino}  &    \cellcolor{lightgrey}74.3      &   \cellcolor{lightgrey}75.6    &  \cellcolor{lightgrey} 75.6   &   \cellcolor{lightgrey}57.3 &  \cellcolor{lightgrey}54.7  & \cellcolor{lightgrey}54.9  &  \cellcolor{lightgrey}50.6     & \cellcolor{lightgrey} 51.6  &  \cellcolor{lightgrey}51.6     \\
\midrule
BEVDepth \cite{li2023BEVDepth}    &  62.9   &   64.8       &   64.2    &   41.5    &  44.8      &   44.9       &   32.9    &   33.1   & 33.0   \\
{\bf Pro3D} w/ BEVDepth &   68.0       &   68.3    &   68.9    &  45.9      &   47.7       &   47.6    &   42.1  &  42.6  & 42.6   \\
\midrule
BEVHeight \cite{yang2023BEVHeight}   &   63.5       &   64.7    &   64.5    &   41.4  &  44.6  & 44.7  &  33.6      &   33.7   &   33.9\\
{\bf Pro3D} w/ BEVHeight &    68.1       &   68.8    &   68.8       &   47.9    &   49.5  &  49.4  & 42.8    &  43.5      &   43.5   \\
\midrule
BEVSpread \cite{wang2024bevspread}   &   64.1       &   65.2    &   64.6    &   41.7  &  45.1  & 44.9  &  33.9     &   34.1   &   34.3\\
{\bf Pro3D} w/ BEVSpread &    68.7       &   69.4    &   69.2       &   48.7    &   50.3  &  50.3  & 44.1    &  44.9      &   44.9   \\

\bottomrule
\end{tabular}
}
\vspace{-2mm}
\label{tab:benchmarking-results of 2D bbox}
\end{table}
}

\section{Cross-Dataset Evaluation}
\label{sec:cross-dataset-evaluation}

One may wonder the results of cross-dataset evaluation. 
The datasets (DAIR-V2X-I and Rope3D) used in our work do not have the same vocabulary,
but both have the {\tt vehicle} class. 
Therefore, 
focusing on the {\tt vehicle} class,
we use the models trained on Rope3D and test on the DAIR-V2X-I benchmark.
In this study, we compare our Pro3D that uses the DINO 2D-detector.
The table below compares BEVHeight, BEVSpread, and Pro3D.
Performance of all methods degrade (marked in parentheses) but our Pro3D yields less degradation compared with the other methods.
{
\setlength{\tabcolsep}{3.20mm}
\begin{table}[h]
\small
\centering
\vspace{-3mm}
\scalebox{0.80} {
\begin{tabular}{l lll lll lll}
\toprule
Method on {\tt vehicle} & {\tt Easy}   & {\tt Mid}   & {\tt Hard}   \\
\midrule
BEVHeight \cite{yang2023BEVHeight}      & 62.6 (-15.2) & 50.1 (-15.7) & 50.3 (-15.6) \\
Pro3D w/ BEVHeight   & 70.0 (-10.7) & 60.7 (-11.1) & 61.1 (-10.9) \\
\midrule
BEVSpread \cite{wang2024bevspread}       & 65.9 (-13.2) & 53.0 (-13.8) & 53.3 (-13.6) \\
\textbf{Pro3D} w/ BEVSpread   & 77.6 (-7.6) & 65.1 (-8.0) & 65.3 (-8.3) \\
\bottomrule
\end{tabular}
}
\vspace{-4mm}
\label{tab:diff_datasets}
\end{table}
}

\section{Comparison on the Inference Speed}
\label{sec:inference time}
We compare the inference time of different methods in Table~\ref{tab:inference time}. BEVHeight~\cite{yang2023BEVHeight} gets the fastest inference time in previous work, so we use it in our ablation studies. Moreover, our method Pro3D outperforms other works in inference speed, where we use YOLOV7 as the 2D detector in Pro3D.

\section{Open-Source Code}
\label{sec:Demo-code}

{\bf Code}.
We include our self-contained codebase (refer to the zip file {\tt Pro3D-main})
as a part of the supplementary material.
Please refer to {\tt README.md}  for instructions how to use the code.
We do not include model weights in the supplementary material as they are too larger than the space limit (100MB).
We will open source our code and release our trained models to foster research.

{\bf License}.
We release open-source code under the MIT License to foster future research in this field.



{\bf Requirement}.
Running our code requires some common packages.
We installed Python and most packages through Anaconda. A few other packages might not be installed automatically, such as Pandas, torchvision, and PyTorch, which are required to run our code. Below are the versions of Python and PyTorch used in our work: 
\begin{itemize}
\item Python version: 3.8.0 [GCC 7.5.0]
\item PyTorch version: 1.8.1
\end{itemize}





\section{More Visualizations, Demo Videos, and Jupyter Notebook of Code}
\label{sec:more-visualizations}

\begin{figure*}[t]
\centering
\centering
\includegraphics[trim=0cm 0 0cm 0cm, clip, width=0.99\linewidth]{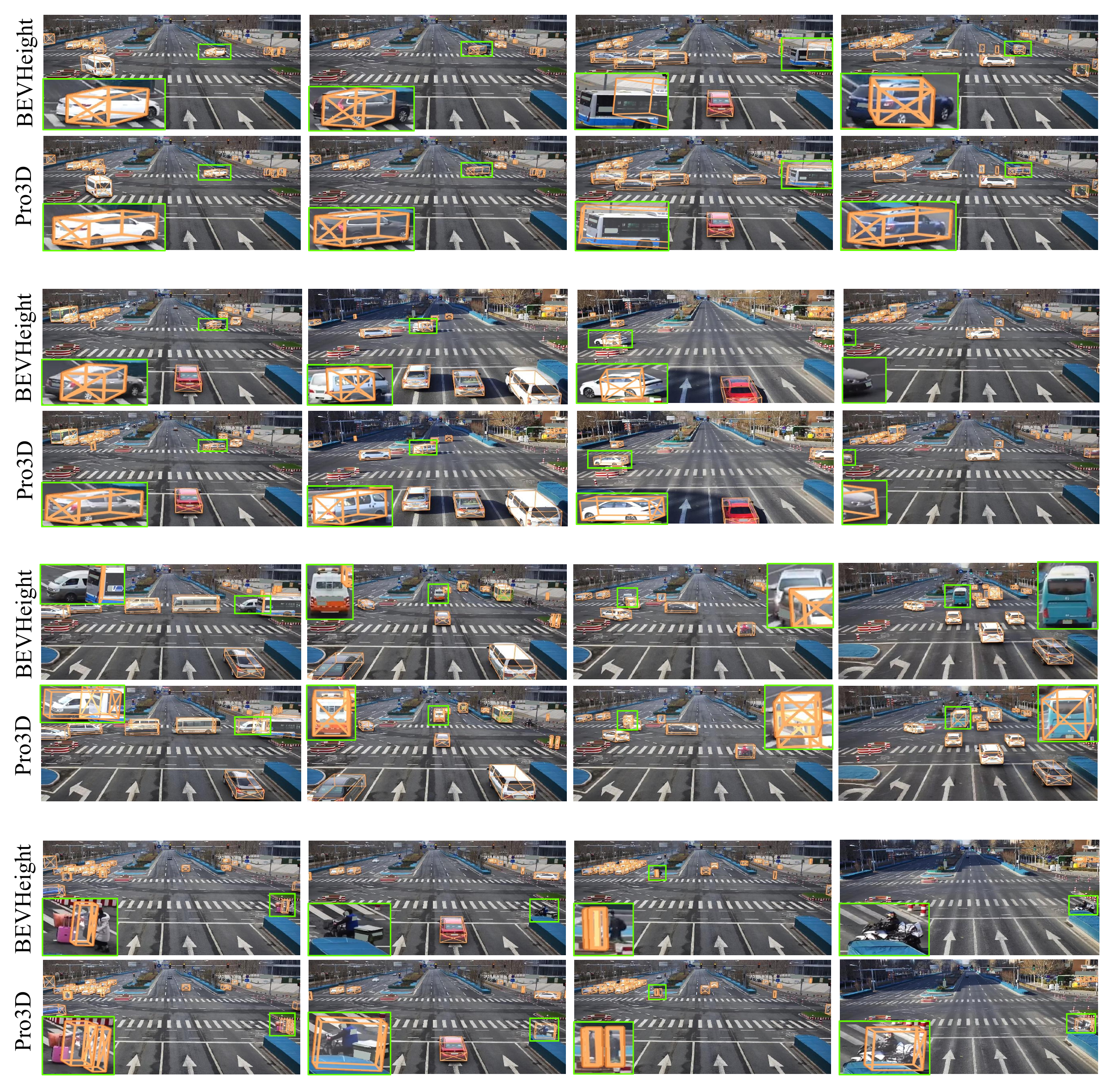}
\vspace{-5mm}
\caption{\small
More visualizations between the state-of-the-art method BEVHeight \cite{yang2023BEVHeight} and our Pro3D.
}
\label{fig:More visualizations of Pro3D}
\vspace{-4mm}
\end{figure*}

Figure~\ref{fig:More visualizations of Pro3D} visualizes more results on the DAIR-V2X-I dataset \cite{yu2022dair}.  
Pro3D not only has better orientation predictions for vehicles but also can detect hard objects such as infrequently-seen ambulances, occluded pedestrians and cyclists. 


{\bf Demo Videos.}
We include two demo videos as a part of the supplementary material, named {\tt demo-video-vs-BEVHeight.mp4} and {\tt demo-video-vs-BEVSpread.mp4}. 
These videos compare our Pro3D with BEVHeight~\cite{yang2023BEVHeight} and BEVSpread~\cite{wang2024bevspread}, respectively. They clearly show that our Pro3D performs much better than both methods in roadside monocular 3D detection.

{\bf Jupyter Notebook of Code.}
In the supplement, we provide not only our code but also a Jupyter Notebook file  named {\tt demo-pro3d-infer-vis.ipynb} which can readily run our Pro3D and display its results on video frame examples. The reader is referred to this file for a quick visualization and guideline on how to run our code.

\end{document}